\documentclass[10pt,journal,compsoc]{IEEEtran}

\usepackage{ragged2e}
\ifCLASSOPTIONcompsoc

\usepackage[nocompress]{cite}
\else

\usepackage{cite}
\fi

\usepackage{url}

\usepackage{times}
\usepackage{graphicx} 
\usepackage{subfigure} 

\usepackage{algorithm}
\usepackage{algorithmic}

\usepackage{hyperref}

\usepackage{helvet}
\usepackage{courier}

\usepackage{makecell}
\usepackage{graphicx}
\usepackage{subfigure}
\usepackage{color}
\usepackage{amsfonts}
\usepackage{amsmath}
\usepackage{amssymb}

\usepackage{multirow}
\usepackage{booktabs}

\def\etal{{\em et al.\/}\, }

\def\mV{{\mathcal V}}

\DeclareMathAlphabet\mathbfcal{OMS}{cmsy}{b}{n}

\def\0{{\bf 0}}
\def\1{{\bf 1}}

\def\bA{{\bf A}}

\def\bW{{\bf W}}
\def\bX{{\bf X}}

\def\bZ{{\bf{Z}}}

\def\bu{{\bf u}}
\def\bv{{\bf v}}

\def\mmR{{\mathbb R}}

\def\bX{{\bf X}}

\def\bW{{\bf W}}

\def\ie{\mbox{\textit{i.e.}}}
\def\eg{\mbox{\textit{e.g.}}}
\def\wrt{\mbox{\textit{w.r.t. }}}

\usepackage{ntheorem}

\newtheorem*{*thm}{Theorem}

\newtheorem*{*lemma}{Lemma}

\usepackage{enumitem}

\def\lzp{\textcolor{cyan}}
\def\guo{\textcolor{blue}}
\def\qi{\textcolor{magenta}}

\def\lzp{\textcolor{black}}
\def\guo{\textcolor{black}}
\def\qi{\textcolor{black}}

\newcommand{\eat}[1]{}
\usepackage{xspace}
\newcommand{\sexyname}{NAT\xspace}
\newcommand{\sexynamei}{NAT\xspace}
\newcommand{\sexynameii}{NAT++\xspace}

\def\mytitle{Towards Accurate and Compact Architectures via Neural Architecture Transformer}

\begin{document}
	
	\title{\mytitle}
	
	\author{Yong Guo$^*$, Yin Zheng$^*$, Mingkui Tan$^*$$^\dagger$, Qi Chen, Zhipeng Li, Jian Chen$^\dagger$, Peilin Zhao, Junzhou Huang
	\IEEEcompsocitemizethanks{
	\IEEEcompsocthanksitem{Yong Guo, Mingkui Tan, Qi Chen, Zhipeng Li, and Jian Chen are with the School of Software Engineering, South China University of Technology. E-mail: \{guo.yong, sechenqi, sezhipengli\}@mail.scut.edu.cn \{mingkuitan, ellachen\}@scut.edu.cn}
	\IEEEcompsocthanksitem{Yin Zheng is with the Weixin Group, Tencent. E-mail: yzheng3xg@gmail.com}
	\IEEEcompsocthanksitem{Peilin Zhao and Junzhou Huang are with Tencent AI Lab, China. E-mail: \{masonzhao, joehhuang\}@tencent.com}
	}
    \thanks{$^*$ Authors contributed equally. $^\dagger$ Corresponding author.}
    }

	\markboth{Journal of \LaTeX\ Class Files, 2021}%
	{Shell \MakeLowercase{\textit{et al.}}: \mytitle}
	
	\IEEEtitleabstractindextext{%
		\begin{abstract}
		\justifying
            Designing effective architectures is one of the key factors behind the success of deep neural networks. Existing deep architectures are either manually designed or automatically searched by some {Neural Architecture Search} (NAS) methods. However, even a well-{designed/}searched architecture may still contain many {nonsignificant} or redundant modules/operations (\eg, {some intermediate} convolution or pooling {layers}). 
            {Such redundancy}
            may not only incur substantial memory consumption and computational cost but  also deteriorate the performance. Thus, it is necessary to optimize the operations inside an architecture to improve the performance without introducing extra computational cost. 
            To this end, we have proposed a Neural Architecture Transformer (\sexyname) method
            which casts the optimization problem into a Markov Decision Process (MDP) and
            seeks to replace the redundant operations with more efficient operations, such as skip or null connection.
            Note that \sexyname only considers a small number of possible transitions and thus comes with a limited search/transition space.
            As a result, such a small search space may hamper the performance of architecture optimization.
            To address this issue, we propose a Neural Architecture Transformer++ (\sexynameii) method which further enlarges the set of candidate transitions to improve the performance of architecture optimization. Specifically, 
            we \guo{present a two-level transition rule to obtain valid transitions, \ie, allowing operations to have more efficient types (\eg, convolution${\to}$separable convolution) or smaller kernel sizes (\eg, $5{\times}5 {\to} 3{\times}3$).} 
            \guo{
            Note that different operations may have different valid transitions.
            We further propose a Binary-Masked Softmax (BMSoftmax) layer to omit the possible invalid transitions.}
            Last, based on the MDP formulation of  NAT and NAT++, we apply policy gradient to learn an optimal policy, which will be used to infer the optimized architectures. 
            We apply \sexyname and \sexynameii to optimize both hand-crafted architectures and NAS based architectures. Extensive experiments on \lzp{several benchmark datasets}
            show that the transformed architecture significantly outperforms both its original counterpart and the architectures optimized by existing  methods.
		\end{abstract}
		
		}

	\maketitle

	\IEEEdisplaynontitleabstractindextext

	\IEEEpeerreviewmaketitle

	{\section{Introduction}\label{sec:introduction}}
	
	\IEEEPARstart{D}{eep} neural networks (DNNs)~\cite{lecun1989backpropagation} have {produced} state-of-the-art results in many challenging tasks including image classification~\cite{guo2018double,krizhevsky2012imagenet,srivastava2015training,Jiang:2017:VDE:3172077.3172161,wei2015hcp,rao2018runtime}, face recognition~\cite{schroff2015facenet,sun2015deeply,ranjan2017hyperface}, 
	and object detection~\cite{ren2016faster,ren2015object,zhang2015accelerating}.
	One of the key factors behind the success lies in the innovation of neural architectures, such as VGG~\cite{simonyan2014very} and
	ResNet\cite{he2016deep}.
	However, designing {effective neural architectures is often {very} labor-intensive and relies heavily on human expertise.} 
	{More critically, {such} a human-designed process {is hard to} fully explore the whole architecture {design} space. {As a result}, the resultant architectures {are often very redundant and} may not be optimal.}
	Hence, there is a growing interest {in replacing} {the manual process of architecture design with {an automatic way called} {Neural Architecture Search} (NAS).}

	Recently, substantial studies~\cite{liu2018darts,pham2018efficient,zoph2016neural,guo2020breaking} have shown that automatically discovered architectures are able to achieve highly competitive performance compared to {the} hand-crafted architectures. However, there are some limitations {to the NAS based} architecture design methods. In fact, since there is an extremely large search space~\cite{pham2018efficient,zoph2016neural} (\eg, billions of candidate architectures), these methods {are hard to be trained and} often produce sub-optimal architectures, leading to the limited representation performance or {substantial computational cost.}
	{Thus, even for {the architectures searched by NAS methods}, it is {still} necessary to optimize {their redundant operations} to achieve better performance and/or reduce the computational cost.}
	
\begin{figure*}[t]
		\centering
		\includegraphics[width=1.9\columnwidth]{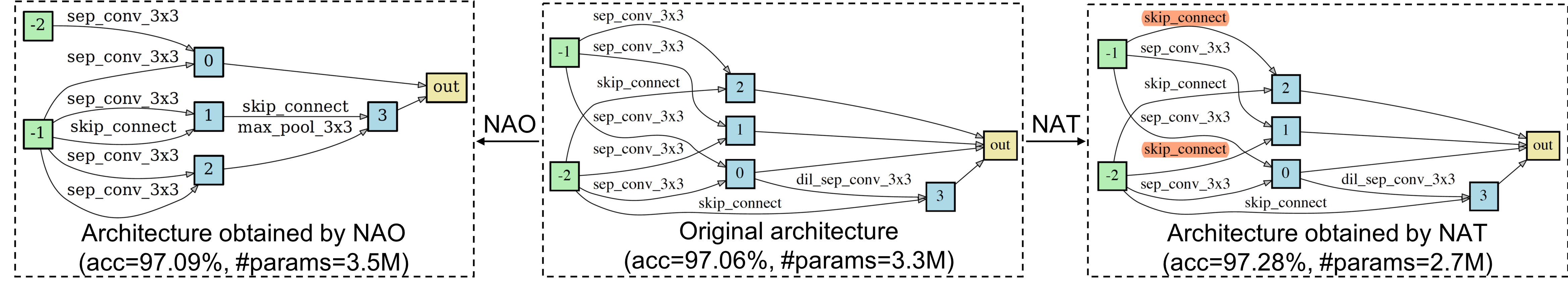}
		\caption{Comparison between {Neural Architecture Optimization} (NAO)~\cite{luo2018neural} and our {Neural Architecture Transformer} (\sexyname). Green blocks denote the two input nodes of the cell and blue blocks denote the intermediate nodes. Red blocks denote the connections that are changed by \sexyname. The accuracy and the number of parameters are evaluated on CIFAR-10 models. 
		}
		\label{fig:example}
	\end{figure*}

	To optimize the architectures, Luo \emph{et al.} recently  proposed a Neural Architecture Optimization (NAO) method~\cite{luo2018neural}. {Specifically}, NAO first encodes an architecture into an embedding in the continuous space and then conducts gradient descent to obtain a better embedding. After that, it uses a decoder to {map the embedding back to obtain an optimized architecture}. However, NAO {has} its own set of limitations. First, NAO often produces a totally different architecture from the input {architecture}, {making it hard to analyze the relationship between the optimized and the original architectures (See Fig.~\ref{fig:example}).} {Second, NAO} may {improve the architecture design at the expense of} introducing extra parameters or {computational} cost. {Third}, similar to the NAS methods, NAO has a {very large} search space, which may not be necessary for architecture optimization and may make the optimization problem very expensive to solve. An illustrative comparison between our methods and NAO can be found in Fig.~\ref{fig:example}. 

	Unlike existing methods that design/find neural architectures,
	we have proposed a Neural Architecture Transformer (\sexyname)~\cite{guo2019nat} method
	to automatically optimize neural architectures to achieve better performance and/or lower computational cost.
	To this end, \sexyname replaces the expensive operations or redundant modules in an architecture with the more efficient operations. 
	{Note} that \sexyname can be used as a general architecture optimizer {that} takes any architecture as input and outputs an optimized {architecture}.
	\sexyname has shown great performance in optimizing various architectures on \lzp{several benchmark datasets}.
	\guo{However, \sexyname only considers three operation transitions, \ie, remaining unchanged, replacing with null connection, replacing with skip connection. 
	Such a small search/transition space may hamper the performance of architecture optimization.
	Thus, it is important and necessary to enlarge the search space of architecture optimization.}

    \guo{In this paper, based on \sexynamei, we propose a Neural Architecture Transformer++ (\sexynameii) method which considers {a larger search space to conduct architecture optimization in a finer manner}. 
    To this end, we present a two-level transition rule to simultaneously change both the type and the kernel size of an operation in architecture optimization. 
    Specifically, \sexynameii encourages operations to have more efficient types (\eg, convolution${\to}$separable convolution) or smaller kernel sizes (\eg, $5{\times}5 {\to} 3{\times}3$).
    For convenience, we use \emph{valid transitions} to denote those transitions that do not increase the computational cost.
    Note that different operations may have different valid transitions.
    To make \sexynameii accommodate all the considered operations, we propose a Binary-Masked Softmax (BMSoftmax) layer to omit all the invalid transitions that violate the transition rule.
    In this way, \sexynameii is able to predict the optimal transitions for the operations with different valid transitions simultaneously.
    Extensive experiments show that our \sexynameii significantly outperforms existing methods.
    }
	
	The contributions of this paper are summarized as follows.

	\begin{itemize}[leftmargin=*]
		
		\item We propose a Neural Architecture Transformer (\sexynamei) method which optimizes arbitrary architectures
		for better performance and/or less computational cost. 
		To this end, \sexynamei either removes the redundant operations or replaces them with skip connections. 
    To better exploit the adjacency information of operations in an architecture, we propose to exploit graph convolutional network (GCN) to build the architecture optimization model.
	
		
		\item Based on NAT, we propose a Neural Architecture Transformer++ (\sexynameii) method which considers a larger search space for architecture optimization. Specifically, \sexynameii presents a two-level transition rule which encourages operations to have a more efficient type and/or a smaller kernel size. 
		Thus, \sexynameii is able to automatically obtain the valid transitions (\ie, the transitions to more efficient operations). 
		
		\item 
		\guo{To accommodate the operations which may have different valid transitions,
		we propose a Binary-Masked Softmax (BMSoftmax) layer to build a general \sexynameii model which predicts the optimal transitions for all the operations simultaneously.}
		
		\item {Extensive experiments on several benchmark datasets show that our \sexynamei and \sexynameii consistently improve the design of various architectures, including both hand-crafted and NAS based architectures. Compared to the original architectures, the optimized architectures tend to yield significantly better performance and/or lower computational cost.}
	
	\end{itemize}

{This paper extends our preliminary version~\cite{guo2019nat} from several aspects. 1) We propose an advanced version \sexynameii by enlarging the search space to improve the performance of architecture optimization. 
\guo{2) We present a two-level transition rule to automatically obtain the valid transitions for each operation on both the operation type level and the kernel size level. 3) We propose a Binary-Masked Softmax (BMSoftmax) layer to omit all the invalid transitions.}
4) We compare the computational cost of different operations and analyze the effect of the transitions among them on our method. 
\guo{5) We provide more analysis about the impact of different operations on the convergence speed of architectures.
6) We investigate the possible bias towards the architectures with too many skip connections in the proposed method.}
7) We provide more empirical results
to show the effectiveness of \sexyname and \sexynameii based on various architectures. 
}

	\begin{figure*}[t]
		\centering
		\subfigure[]{
			\includegraphics[width=0.45\columnwidth]{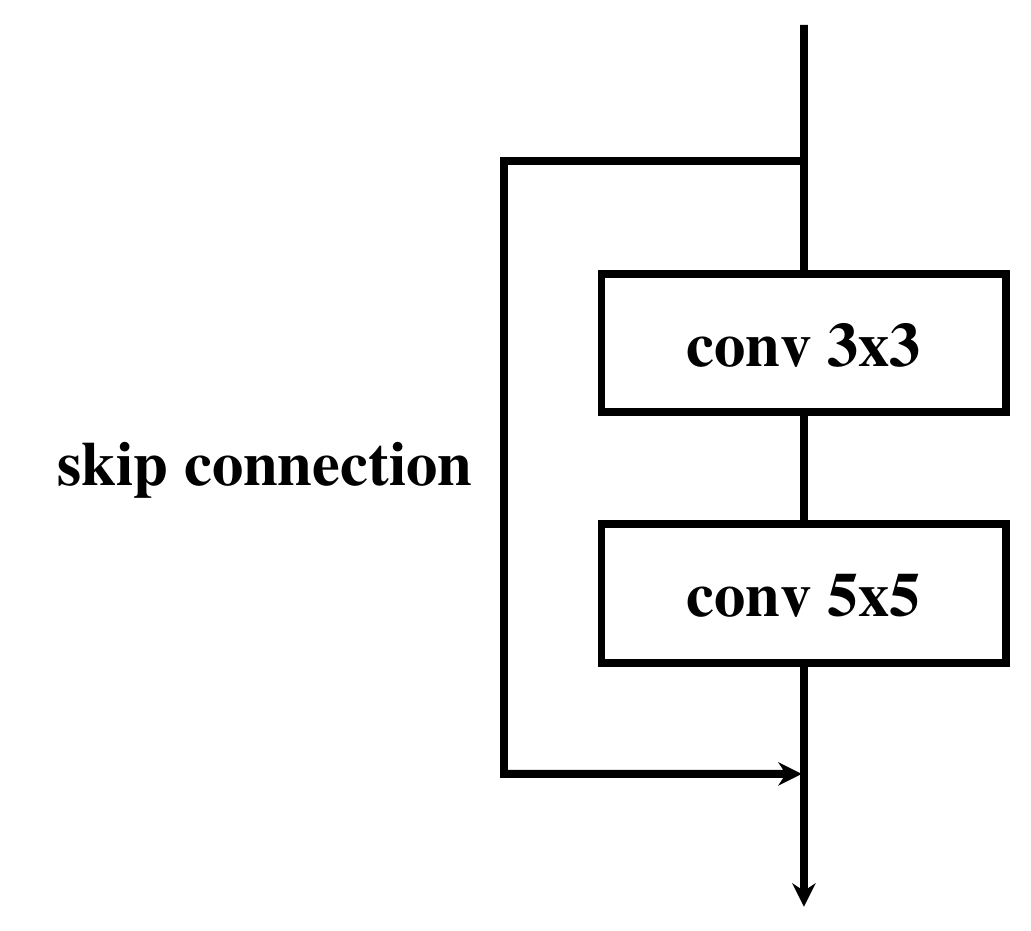}\label{fig:eg_orig_arch}
		}
		\hspace{0.1\columnwidth}
		\subfigure[]{
			\includegraphics[width=0.45\columnwidth]{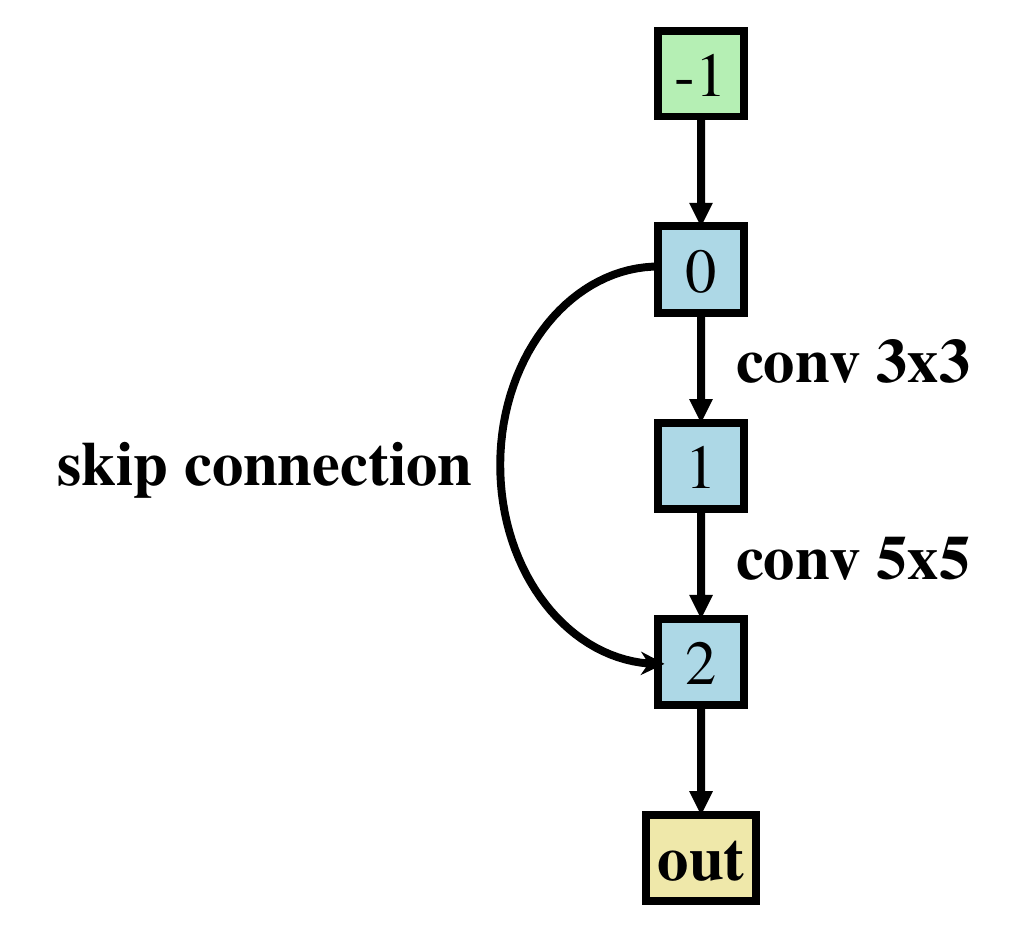}\label{fig:eg_pruned_arch}
		}
		\hspace{0.1\columnwidth}
		\subfigure[]{
			\includegraphics[width=0.45\columnwidth]{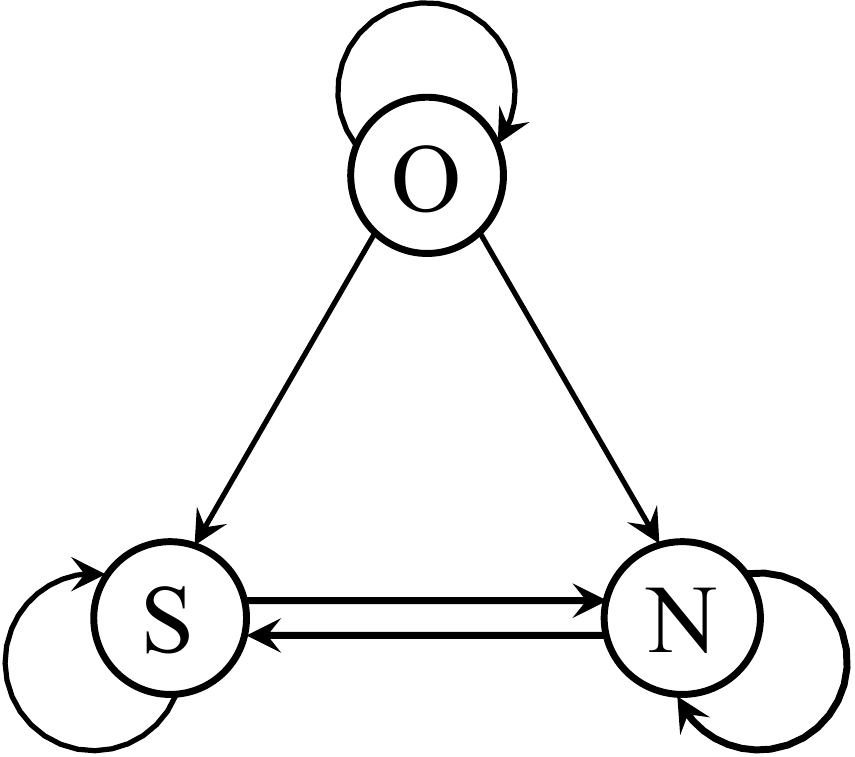}\label{fig:state_transition}
		}
		\caption{An example of the {graph representation of a residual block and the diagram of operation transformations.} (a) a residual block~\cite{he2016deep}; (b) a graph view of residual block; (c) {transitions among three kinds of operations.} ${N}$ denotes a null operation without any computation, ${S}$ denotes a skip connection, and ${O}$ denotes some computational modules other than null and skip connections.}
		\label{fig:arch2graph}
	\end{figure*}

	\section{Related Work} \label{sec:related_work}
	
	\subsection{Hand-crafted Architecture Design}
	Many studies have 
	{proposed} a series of deep neural architectures, such as AlexNet~\cite{krizhevsky2012imagenet},
	VGG~\cite{simonyan2014very} and so on. 
    Based on these models, {many efforts have been made to further increase the representation ability of deep networks. Szegedy \emph{et al.} propose the GoogLeNet~\cite{szegedy2015going} which consists of a set of convolutions with different kernel sizes.} He \emph{et al.} propose the residual network (ResNet)~\cite{he2016deep} by introducing residual shortcuts between different layers. 
    To design more compact models, 
	MobileNet~\cite{howard2017mobilenets,sandler2018mobilenetv2} employs depthwise separable convolution to reduce model size and computational overhead.
	{ShuffleNet~\cite{zhang2018shufflenet,ma2018shufflenet} exploits pointwise group convolution and channel shuffle to significantly reduce computational cost while maintaining comparable accuracy. 
	}
	However, the human-designed process often requires substantial human effort and cannot fully explore the whole architecture space.

	\subsection{Neural Architecture Search}
	{Recently,} neural architecture search (NAS) has been proposed
	to automate the process of architecture design~\cite{baker2016designing, zhong2018practical,cai2018proxylessnas,vaswani2017attention,so2019evolved}.
	Specifically, Zoph~\etal use a recurrent neural network as the controller~\cite{zoph2016neural} to construct each convolution by determining the optimal stride,  the number and the shape of filters.
	Pham~\etal propose a weight sharing technique~\cite{pham2018efficient} to significantly improve search efficiency.
	Liu~\etal propose a differentiable NAS method, called DARTS~\cite{liu2018darts}, which relaxes the search space to be continuous. 
	Recently, 
	Luo \emph{et al.} propose the Neural Architecture Optimization (NAO)~\cite{luo2018neural} method to perform architecture search on continuous space by exploiting encoding-decoding technique. 
	Unlike these methods, our method optimizes architectures without introducing extra computational cost (See comparisons in Fig.~\ref{fig:example}).

	\subsection{Architecture Adaptation and Model Compression}
	Several methods have been proposed to {adapt architectures to some specific platform or compress some existing architectures.}
	To obtain compact models, ~\cite{yang2018netadapt,lemaire2019structured,dai2019chamnet,chen2015net2net} adapt architectures to the more compact ones by learning the optimal settings of each convolution.
	One can also exploit model compression methods~\cite{li2016pruning,he2017channel,luo2017thinet,zhuang2018discrimination} to remove the redundant channels to obtain compact models. 
	\lzp{Recently, ESNAC~\cite{cao2019learnable} uses Bayesian optimization techniques to search for a compressed network via layer removal, layer shrinkage, and adding skip connections. ASP~\cite{wang2020revisiting} proposes an affine parameter sharing method to search for the optimal channel numbers of each layer to optimize architectures.} Nevertheless, these methods have to learn a compressed model for a specific architecture and have limited generalization ability to different architectures.
	Unlike these methods, we seek to learn a general optimizer for arbitrary architecture.

	\begin{figure*}[t]
		\centering
		\includegraphics[width=1.9\columnwidth]{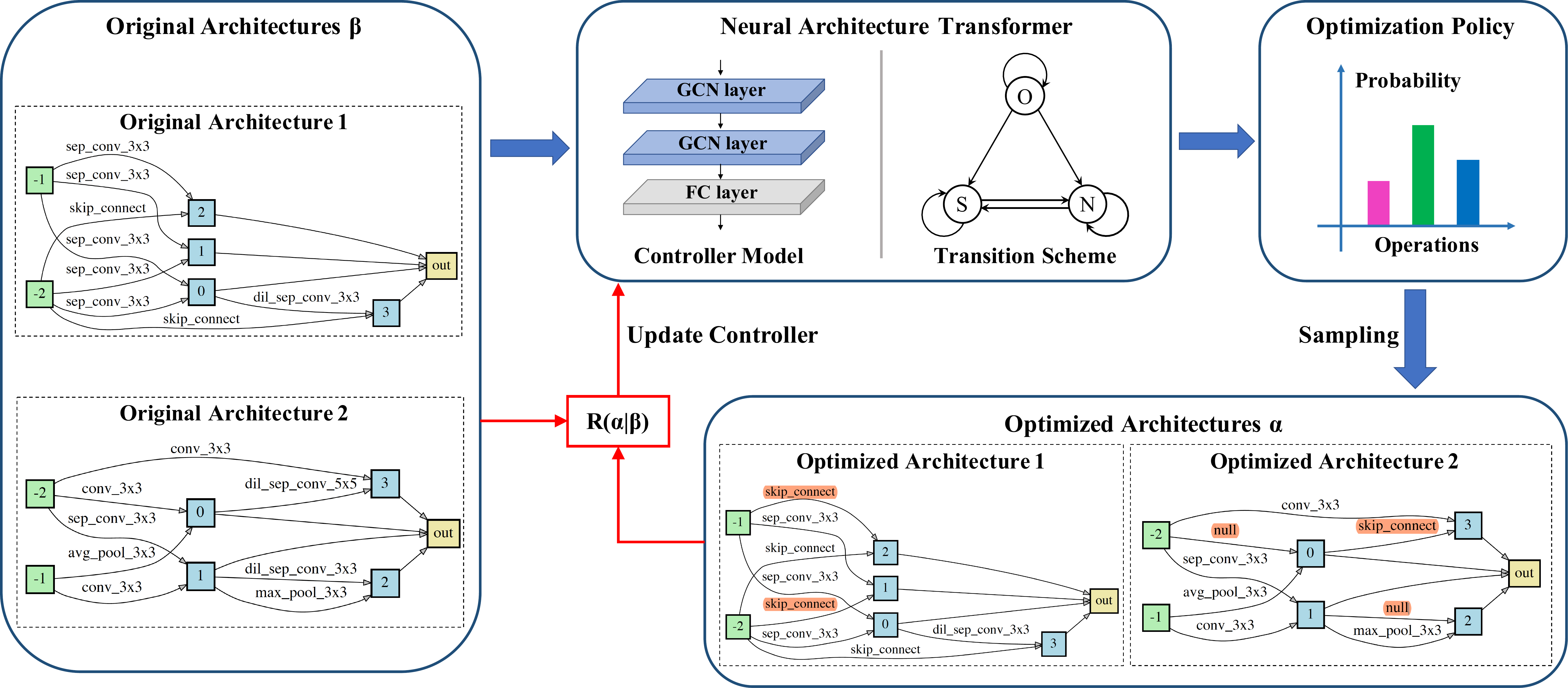}
		\caption{{The scheme of the proposed \sexyname. Our \sexyname takes an arbitrary architecture as input and produces the optimized architecture as the output. We use blue arrows to represent the process of architecture optimization. Red arrows and boxes denote the computation of reward and gradients. $R(\alpha | \beta)$ denotes the reward that measures the performance improvement between two architectures $\alpha$ and $\beta$.}}
		\label{fig:nat_scheme}
	\end{figure*}

	\section{Neural Architecture Transformer}\label{sec:NAT}


	\subsection{Problem Definition}\label{sec:problem_definition}
	
	Following~\cite{pham2018efficient,liu2018darts}, we consider a cell as the basic block to build the entire network.
	Given a cell based architecture space $\Omega$, we can represent an architecture $\alpha$ as a directed acyclic graph (DAG), \ie, $\alpha=\left(\mathcal{V},\mathcal{E}\right)$, where ${\mathcal{V}}$ is a set of nodes that denote the feature maps in DNNs and $\mathcal{E}$ is an edge set~\cite{zoph2016neural,pham2018efficient,liu2018darts}, as shown in Fig.~\ref{fig:arch2graph}. 
	Here,
	\qi{a DAG contains $|\mV|$ nodes $\{\bX_l\}_{l=-2}^{|\mV|-3}$, where $\bX_{-2}$ and $\bX_{-1}$ denote the outputs of two previous cells, and $\bX_{|\mV|-3}$ denotes the output node that concatenates all intermediate nodes $\{\bX_l\}_{l=0}^{|\mV|-3}$. Each intermediate node is able to connect with all previous nodes.}
	The directed edge $e_{ij} \in {\mathcal{E}}$ denotes some operation (\textit{\eg,} convolution or max pooling) that transforms the feature map from node $v_i$ to $v_j$. 
	For convenience, we divide the edges in ${\mathcal{E}}$ into three categories, namely, $S$, ${N}$, $O$, as shown in Fig.~\ref{fig:state_transition}.  Here, ${S}$ denotes the skip connection, ${N}$ denotes the {null connection} (\ie, no edge between two nodes), and ${O}$ denotes the operations other than skip connection or null connection (\eg, convolution or max pooling). Note that different operations have different cost. Specifically, let $c(\cdot)$ be a function to evaluate the computational cost. Obviously, we have $c({O}) > c({S}) > c({N})$.

	In this paper, we propose 
	an architecture optimization method, called Neural Architecture Transformer (NAT),
	to optimize any given architecture to achieve better performance and/or less computational cost. 
	To avoid introducing extra computational cost, an intuitive way is to make the original operation have less computational cost, \eg, replacing operations with skip or null connection.
	Although skip connection has a slightly higher cost than null connection, it often can significantly improve the performance~\cite{he2016deep,he2016identity}. 
	Thus, we enable the transition from null connection to skip connection to increase the representation ability of deep networks. In summary, we constrain the possible transitions among $O$, $S$ and ${N}$ in Fig.~\ref{fig:state_transition} in order to reduce the computational cost.


	\subsection{Markov Decision Process for \sexyname}
	\label{sec:MDP}

	In this paper, we seek to learn a general architecture optimizer which takes any given architecture as input and outputs the corresponding optimized architecture. Let $\beta$ be the input architecture which follows some distribution $p(\cdot)$, \eg, multivariate uniformly discrete distribution.
	We seek to obtain the optimized architecture $\alpha$ by learning the mapping $\alpha \leftarrow {\rm NAT} (\beta; \theta)$, where $\theta$ denotes the learnable parameters.
	{Let $w_{\alpha}$ and $w_{\beta}$ be the well-learned model parameters of architectures $\alpha$ and $\beta$, respectively.} We measure the performance of $\alpha$ and $\beta$ by some metric $R(\alpha, w_{\alpha})$ and $R(\beta, w_{\beta})$, \eg, accuracy. 
	For convenience, we define the performance improvement between $\alpha$ and $\beta$ by $R(\alpha | \beta) = R(\alpha, w_{\alpha}) - R(\beta, w_{\beta})$. 
    To illustrate our method, we first discuss the architecture optimization problem for a specific architecture and then generalize it to the problem for different architectures.

    To learn a good architecture transformer ${\rm NAT} (\beta; \theta)$ to optimize a specific $\beta$, we can maximize the performance improvement $R(\alpha | \beta)$.
	However, simply maximizing $R(\alpha | \beta)$ may easily find an architecture $\alpha$ with much higher computational cost than the input counterpart $\beta$. 
	Instead, we seek to obtain the optimized architectures with better performance without introducing additional computational cost.
	To this end, we introduce a constraint $c(\alpha) \leq c(\beta)$ to encourage the optimized architecture to have lower computational cost than the input one. 
Moreover, it is worth mentioning that, directly obtaining the optimal $\alpha$ \wrt the input architecture $\beta$ is non-trivial~\cite{zoph2016neural}.
Following~\cite{zoph2016neural,pham2018efficient},
we instead learn a policy $\pi(\cdot | \beta; \theta)$ and use it to produce an optimized architecture, \ie, $\alpha {\sim} \pi(\cdot|\beta; \theta)$.
To learn the policy, we seek to solve the following optimization problem:
\begin{equation}\label{eq:objective}
        	\begin{aligned}
        	&\max_{\theta} ~  \mathbb{E}_{\alpha \sim \pi(\cdot|\beta; \theta)} ~[R \left( \alpha | \beta \right)],\\
        	& ~\text{s.t. } ~c(\alpha) \leq c(\beta), ~\alpha \sim \pi(\cdot|\beta; \theta),
        	\end{aligned}
\end{equation}
	{where $\mathbb{E}_{\alpha \sim \pi(\cdot|\beta; \theta)} \left[  \cdot \right]$ denotes the expectation operation over $\alpha$.}

However, the optimization problem in Eqn.~(\ref{eq:objective}) only focuses on a single input architecture. To learn a general architecture transformer that is able to optimize any given architecture, we maximize the expectation of performance improvement $R(\alpha | \beta)$ over the distribution of input architecture $\beta {\sim} p(\cdot)$. Formally, the expected performance improvement over different input architectures can be formulated by $\mathbb{E}_{\beta \sim p(\cdot)} \left[  \mathbb{E}_{\alpha \sim \pi(\cdot|\beta; \theta)} ~R \left( \alpha | \beta \right) \right]$.
Consequently, the optimization problem becomes
\begin{equation}\label{eq:objective-final}
        	\begin{aligned}
        	&\max_{\theta} ~\mathbb{E}_{\beta \sim p(\cdot)} \left[  \mathbb{E}_{\alpha \sim \pi(\cdot|\beta; \theta)} ~R \left( \alpha | \beta \right) \right],\\
        	& ~\text{s.t. } ~c(\alpha) \leq c(\beta), ~\alpha \sim \pi(\cdot|\beta; \theta).
        	\end{aligned}
\end{equation}

    Unlike conventional neural architecture search (NAS) methods that design/find an architecture from scratch~\cite{pham2018efficient,liu2018darts}, we hope to optimize any given architectures
    by replacing redundant operations (\eg, convolution) in the input architecture with the more efficient ones (\eg, skip connection).
    Since we only allow the transitions that do not increase the computational cost (also called \emph{valid transitions}) in Fig.~\ref{fig:state_transition}, compared to the input architecture $\beta$, the optimized architecture $\alpha$ would have less or at least the same computational cost.
    Thus, the proposed method can naturally satisfy the cost constraint $c(\alpha) \leq c(\beta)$.
    
As mentioned above, our \sexyname only takes a single architecture $\beta$ as input to predict the optimized architectures.
However, one may obtain a better optimized architecture if we consider the previous success and failure optimization results/records of other architectures. In this case, the optimization problem would be extremely complicated and hard to solve.
\guo{To alleviate the training difficulty of the optimization problem, we formulate it as a Markov Decision Process (MDP).
Specifically, we exploit the Markov property to optimize the current architecture without considering the previous optimization results (similar to the MDP formulation in the multi-arm bandit problem~\cite{vermorel2005multi,anantharam1987asymptotically}).
In this way, MDP is able to greatly simplify the decision process.
We put more discussions on our MDP formulation in the supplementary.
}

	\textbf{MDP formulation details.} 
	A typical MDP~\cite{schulman2015trust} is defined by a tuple $(\mathcal{S}, \mathcal{A}, P, {R}, q, \gamma)$, 
	where $\mathcal{S}$ is a finite set of states, $\mathcal{A}$ is a finite set of actions, $P: \mathcal{S} \times \mathcal{A} \times \mathcal{S} \rightarrow \mathbb{R}$ is the state transition distribution, ${R}: \mathcal{S} \times \mathcal{A} \rightarrow \mathbb{R}$ is the reward function, $q: \mathcal{S} \rightarrow [0, 1]$ is the distribution of initial state, and $\gamma \in [0, 1]$ is a discount factor. 
	Here, we define an architecture as a state, a transformation mapping $\beta \to \alpha$ as an action. 
	Here, we use the accuracy improvement on the validation set as the reward. Since the problem is {a} one-step MDP, {we can omit the discount factor $\gamma$.}
	Based on the problem definition, {we transform any $\beta$ into an optimized architecture $\alpha$} with the policy $\pi(\cdot|\beta;\theta)$.

	\subsection{Policy Learning by Graph Convolutional Network}\label{sec:graph_representation}
	
	As mentioned in Section~\ref{sec:MDP},
	\sexyname takes an architecture graph $\beta$ as input and outputs the optimization policy $\pi(\cdot|\beta;\theta)$. 
	To learn the optimal policy,
	since the optimization of an operation/edge in the architecture graph depends on the adjacent nodes and edges,
	we consider both the current edge and its neighbors. Therefore, we build the controller model with a graph convolutional networks (GCN)~\cite{kipf2016semi} to exploit the adjacency information of the operations in the architecture.
	Here, an architecture graph can be represented by a data pair $(\bA, \bX)$,
	where $\bA$ denotes the adjacency matrix of the graph and $\bX$ denotes the attributes of the nodes together with their two input edges.
	\qi{We put more details in the supplementary.}

    Note that a graph convolutional layer is able to extract features by aggregating the information from the neighbors of each node (\ie, one-hop neighbors)~\cite{wu2020comprehensive}. Nevertheless, 
    building the model with too many graph convolutional layers (\ie, high-order model) may introduce redundant information~\cite{zhu2019multi} and hamper the performance (See results in Fig.~\ref{fig:num_gcn}).
    In practice, we build our \sexynamei with a two-layer GCN, which can be formulated as
	\begin{equation}\label{eq:gcn_classifier}
	\small
	\bZ = f(\bX, \bA) = h\left(\bA {\rm \sigma}\left(\bA \bX \bW^{(0)}\right)\bW^{(1)}\bW^{\rm FC}\right),
	\end{equation}
	where $\bW^{(0)}$ and $\bW^{(1)}$ denote the weights of two graph convolution layers, $\bW^{\rm FC}$ denotes the weight of the fully-connected layer, $\sigma$ is a non-linear activation function (\eg, ReLU~\cite{nair2010rectified}), $h(\cdot)$ denotes the softmax layer, and $\bZ$ refers to the probability distribution of $\pi(\cdot|\beta; \theta)$ over 3 transitions on the edges, \ie, ``remaining unchanged'', ``replacing with null connection'', and ``replacing with skip connection''.
	\guo{It is worth mentioning that, the controller model is essentially a 3-class GCN based classifier. Given $K$ edges in an architecture, \sexyname outputs the  probability  distribution $\bZ \in \mmR^{K \times 3}$.}
	For convenience, we denote $\theta= \{ \bW^{(0)}, \bW^{(1)}, \bW^{\rm FC} \}$ as the parameters of the controller model of \sexyname.

	\begin{algorithm}[t]
		\small
		\caption{Training method for \sexyname.}
		\label{alg:training}
		\begin{algorithmic}[1]
			\REQUIRE  The number of sampled input architectures in an iteration $m$, the number of sampled optimized architectures for each input architecture $n$, learning rate $\eta$, regularizer parameter $\lambda$ in Eqn.~(\ref{eq:obj_entropy}), input architecture distribution $p(\cdot)$. \\
            \STATE Initialize the parameters $\theta$ and $w$. \\
			\WHILE{not convergent}
			\FOR{each iteration on training data} 
			\STATE // {Fix $\theta$ and update $w$.} \\
			\STATE Sample $ \beta_i \sim p(\cdot)$ to construct a batch $\{\beta_i \}_{i=1}^m$.\\
			\STATE Update supernet parameters $w$ by descending the gradient:\\
			\STATE ~ \vspace{-5 pt}
            \begin{equation*}
            w \leftarrow w - \eta \frac{1}{m} \sum_{i=1}^{m} \nabla_{w} \mathcal{L}(\beta_i,w).
			\end{equation*} 
			\vspace{-8 pt}
			\ENDFOR
			\FOR{each iteration on validation data}
			\STATE // {Fix $w$ and update $\theta$.} \\
			\STATE Sample $\beta_i \sim p(\cdot)$ to construct a batch $\{\beta_i \}_{i=1}^m$.\\
			\STATE Obtain $\{\alpha_j\}_{j=1}^n$ according to the policy learned by GCN.\\
			\STATE Update transformer parameters $\theta$ by ascending the gradient:\\
			\STATE~ \vspace{-13 pt}
			\begin{equation*}
		    	\begin{aligned}
    			  \theta \leftarrow &\theta + \eta \frac{1}{mn} \sum_{i=1}^{m} \sum_{j=1}^{n} \Big[\nabla_{\theta} \log \pi(\alpha_j | \beta_i; \theta) R( \alpha_j | \beta_i) \\
    			  + &\lambda \nabla_{\theta} H \big(\pi \left(\cdot|\beta_i; \theta \right) \big) \Big].
			    \end{aligned}
			\end{equation*}
			\vspace{-10 pt}
			\ENDFOR
			\ENDWHILE
		\end{algorithmic}
	\end{algorithm}

	\subsection{Training Method for \sexyname}\label{sec:train}
	
	As shown in Fig.~\ref{fig:nat_scheme}, given an architecture $\beta$ as input, \sexyname outputs the policy/distribution $\pi(\cdot|\beta;\theta)$ over different candidate transitions. 
	Based on $\pi(\cdot|\beta;\theta)$, we conduct sampling to obtain the optimized architecture $\alpha$. After that, we compute the reward $R(\alpha|\beta)$ to guide the search process. To learn \sexyname, we first update the supernet parameters $w$ and then update the architecture transformer parameters $\theta$ in each iteration.
	We show the detailed training procedure in Algorithm~\ref{alg:training}.

	\textbf{Training the parameters of the supernet $w$.}
	Given any $\theta$, we need to update the supernet parameters $w$ based on the training data. To accelerate the training process, we adopt the parameter sharing technique~\cite{pham2018efficient}. 
	Then, we can use the shared parameters $w$ to represent the parameters for different architectures. 
	For any architecture $\beta \sim p(\cdot)$, let $\mathcal{L}(\beta,w)$ be the loss function, \eg, the cross-entropy loss. Then, given any $m$ sampled architectures, the updating rule for $w$ with parameter sharing can be given by $w \leftarrow w - \eta \frac{1}{m} \sum_{i=1}^{m} \nabla_{w} \mathcal{L}(\beta_i,w)$, where $\eta$ is the learning rate.

	\textbf{Training the parameters of the controller model $\theta$.}
	We train the transformer with reinforcement learning (\ie, policy gradient)~\cite{williams1992simple} for several reasons. 
	\guo{\textbf{{First}}, from Eqn.~(\ref{eq:objective-final}), there are no supervision signals (\ie, ``ground-truth'' better architectures) 
	to train the model in a supervised manner. 
	\textbf{{Second}}, the metrics of both accuracy and computational cost are non-differentiable. As a result, the gradient-based methods cannot be directly used for training.} 
	To address these issues, we use reinforcement learning to train our model by maximizing the expected reward over the optimization results of different architectures.
	
	To encourage exploration, we use an entropy
	regularization term in the objective
	to prevent the transformer from converging to a local optimum too quickly~\cite{zoph2018learning}, \eg, selecting the ``original'' option for all the operations.
	The objective can be formulated as
	\begin{equation}\label{eq:obj_entropy}
    	\begin{aligned}
        	&J(\theta) = \mathbb{E}_{\beta \sim p(\cdot)} [ \mathbb{E}_{\alpha \sim \pi(\cdot |\beta;\theta)}  \left[ {R}\left(\alpha | \beta \right)  \right] {+} \lambda H \big(\pi(\cdot|\beta; \theta) \big)  ]\\
        	& = \sum_{\beta} p(\beta) \big[ \sum_\alpha \pi(\alpha | \beta; \theta) \big( {R}\left(\alpha | \beta \right) \big) + \lambda H \big(\pi(\cdot|{\beta}; \theta) \big) \big],
    	\end{aligned}
	\end{equation}
	where $p(\beta)$ is the probability to sample some architecture $\beta$ from the distribution $p(\cdot)$, $\pi(\alpha|\beta;\theta)$ is the probability to sample some architecture $\alpha$ from the distribution $\pi({\cdot|\beta};\theta)$, $H(\cdot)$ evaluates the entropy of the policy, and $\lambda$ controls the strength of the entropy regularization term.
	For each input architecture, we sample $n$ optimized architectures $\{\alpha_j\}_{j=1}^{n}$ from the distribution $\pi(\cdot | \beta; \theta)$ in each iteration.
	Thus, the gradient of Eqn.~(\ref{eq:obj_entropy}) w.r.t. $\theta$ becomes
	\begin{equation}\label{eq:entropy_gradient}
    	\begin{aligned}
        	\nabla_{\theta} J(\theta) &\approx \frac{1}{mn} \sum_{i=1}^{m} \sum_{j=1}^{n} \Big[ \nabla_{\theta} \log  \pi(\alpha_j | \beta_i; \theta) R(\alpha_j | \beta_i) \\ 
        	&+ \lambda \nabla_{\theta} H \big(\pi(\cdot|\beta_i; \theta) \big) \Big].
    	\end{aligned}
	\end{equation}
	The regularization term $H \big(\pi(\cdot|\beta_i; \theta) \big)$ encourages the distribution $\pi(\cdot|\beta; \theta)$ to have high entropy, 
	\ie, high diversity in the decisions on the edges. Thus, the decisions for some operations would be encouraged to choose the ``skip'' or ``null'' operations during training. In this sense, \sexyname is able to explore the whole search space to find the optimal architecture.

\subsection{Inferring the Optimized Architectures} \label{sec:infer}
    After the training process in Algorithm~\ref{alg:training}, we obtain the parameters $\theta$ of the architecture transformer model ${\rm NAT}(\cdot; \theta)$. Based on the \sexyname model, we take any given architecture $\beta$ as input and output the architecture optimization policy $\pi(\cdot|\beta;\theta)$.
	\lzp{Then, we conduct sampling according to the learned policy to obtain the optimized architecture,
	\lzp{\ie, $\alpha {\sim} \pi(\cdot|\beta; \theta)$.}
	Specifically, we predict the optimal transition among three candidate transitions (\ie, ``remaining unchanged'', ``replacing with null connection'', and ``replacing with skip connection'') for each edge in the architecture graph.
	\qi{Note} that the sampling method is not an iterative process and we perform sampling once for each operation/edge.}
	\qi{We} can also obtain the optimized architecture by selecting the operation with the maximum probability, which, however, tends to reach a local optimum and yields worse results than the sampling based method (See results in supplementary).

    \begin{figure*}[t]
		\subfigure[]{
		    	\begin{minipage}[t]{0.53\linewidth}
            	\centering
            	\includegraphics[width=1\columnwidth]{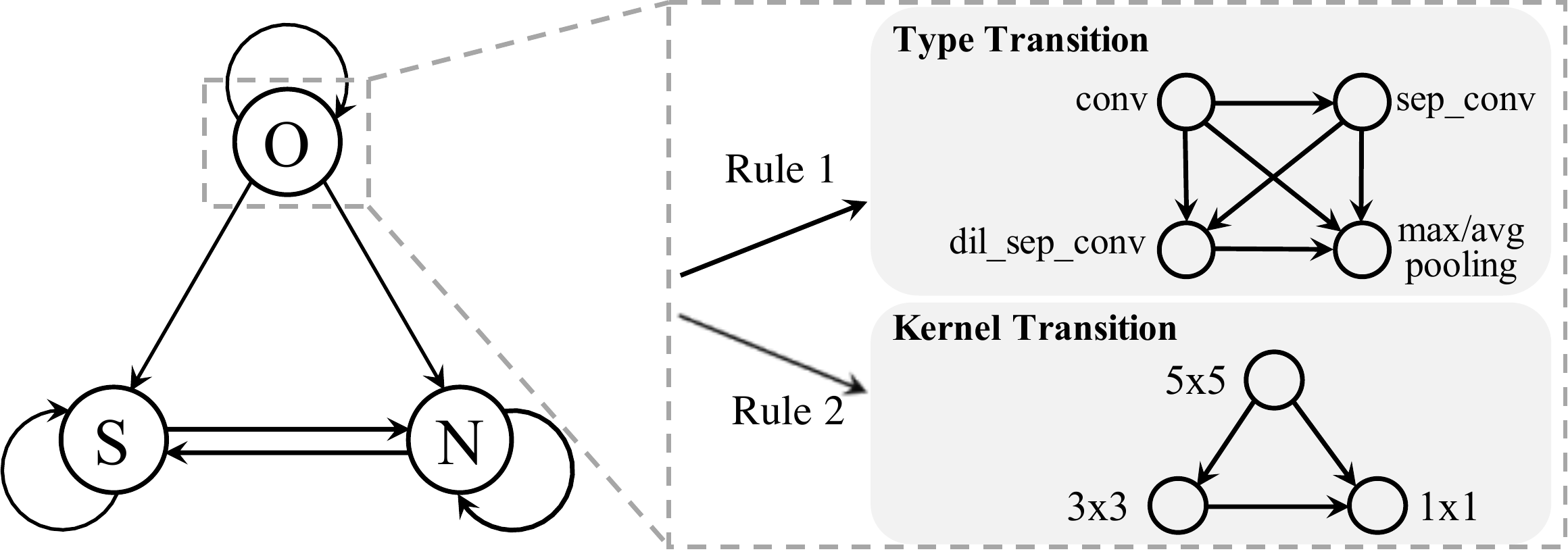}\label{fig:state_transition_natv2}
            	\label{fig:state_transition_natv2}
            	\end{minipage}
		}~~~~~~~
		\subfigure[]{
                \begin{minipage}[t]{0.39\linewidth}
            	\includegraphics[ width=1\linewidth]{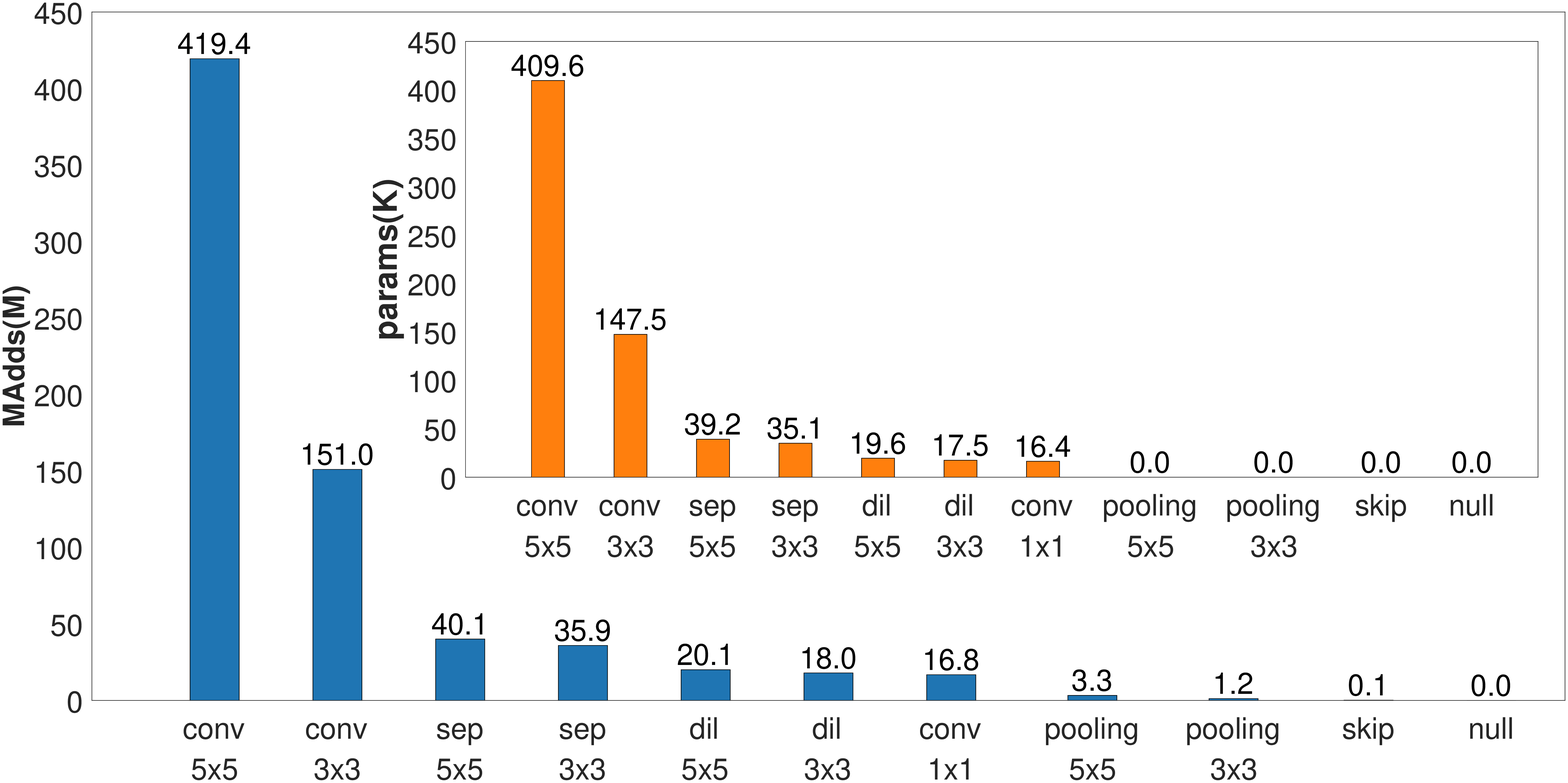}
            	\label{fig:operation_cost}
            	\end{minipage}
		}
		\caption{Operation transition scheme of \sexynameii. (a) Two-level transition rule of \sexynameii; (b) Computational {cost} of different operations. We set the input channel and output channel to 128, the height and width of {the} input feature maps to 32. Here, sep denotes {a} separable convolution and dil denotes {a} {dilated} separable convolution.}
		\label{fig:scheme_natv2}
	\end{figure*}

    \section{Neural Architecture Transformer++}\label{sec:improved_NAT}
	
	As mentioned in Section~\ref{sec:NAT}, \sexynamei replaces the redundant operations in $O$ with the null connections $N$ or the skip connections $S$ according to the transition scheme in Fig.~\ref{fig:state_transition}. \guo{However, there are still several limitations of \sexynamei.
	\textbf{First}, merely replacing an operation with the null or skip connection 
	makes the search space very small and may hamper the performance of architecture optimization.
	\textbf{Second}, when we divide $O$ into more specific operations, the number of transitions between every two categories would significantly increase. As a result, it is non-trivial to manually design valid transitions for each operation using \sexynamei. \textbf{Third}, since operations may have different valid transitions to reduce the computational cost, it is hard to build a general GCN based classifier to predict the optimal transitions for all the operations.
	}
	
	To address the above limitations,
	we further consider more possible operation transitions to enlarge the search space and develop more flexible operation transition rules. The proposed method is called Neural Architecture Transformer++ (NAT++), whose operation transition scheme is shown in Fig.~\ref{fig:scheme_natv2}.
	\guo{
	In \sexynameii, we propose a \textbf{two-level transition rule} 
	which encourages operations to have more efficient types or smaller kernel sizes to produce more compact architectures. 
    Note that different operations may have different valid transitions.
    To predict the optimal transitions for the operations with different valid transitions, we propose a \textbf{Binary-Masked Softmax (BMSoftmax)} layer to build the \sexynameii model. 
    We will depict our \sexynameii in the following.
    }

    \subsection{Operation Transition Scheme for \sexynameii}\label{sec:extended_transition}

    \guo{
    Note that \sexyname~\cite{guo2019nat} only considers three operation transitions, \ie, remaining unchanged, replacing with null connection, replacing with skip connection. As a result, the search space may be very limited and may hamper the performance of architecture optimization.}
    To consider a larger search space, we propose a 
    two-level transition scheme which encourages operations to have more efficient types and/or smaller kernel sizes (See Fig.~\ref{fig:state_transition_natv2}).

    \subsubsection{Two-level Transition Scheme}
    In \sexynameii, we consider a larger search space to enable more possible transitions for architecture optimization.
    Specifically, we allow the transitions among six operation types, namely standard convolution, separable convolution, dilated separable convolution, max/average pooling, skip connection, and null connection. For each operation type, we consider three kernel sizes, \ie, $1{\times}1$, $3{\times}3$, and $5{\times}5$\footnote{We put the details about all the considered operations in supplementary.}.
    To optimize both the type and kernel size of operations, 
    we design a type transition rule and a kernel transition rule, respectively.

    \begin{itemize}
        \item \textbf{Type Transition:} We seek to reduce the computational cost by changing operation into a more computationally efficient one. According to Fig.~\ref{fig:operation_cost}, we use the following rule:
        $$\text{{Rule 1:} conv} \rightarrow \text{sep\_conv} \rightarrow \text{dil\_sep\_conv} \rightarrow \text{pooling},$$
        {where $\rightarrow$ denotes the transition direction. Since max pooling has a similar computational cost to average pooling, we enable the transition between max pooling and average pooling.}
        
        \item \textbf{Kernel Transition:} {Given a specific operation type, one can also adjust the kernel size to change the operation. In general, a larger kernel would induce higher computational cost. Thus, to make sure that all the transitions can reduce the computational cost, we consider the following rule:}
        $$ \text{Rule 2: }  5 \times 5 \rightarrow 3 \times 3 \rightarrow 1 \times 1.$$
        
    \end{itemize}

    {It is worth noting that only using any of the two rules cannot guarantee that we can reduce the computational cost. Specifically, according to Fig.~\ref{fig:operation_cost}, if we only focus on the rule on operation type, there may still exist some transitions that increase the computational cost by changing the operation type to a more efficient one but increasing the kernel size, \eg, conv\_$1{\times}1$ $\to$ sep\_conv\_$3{\times}3$.
    Similarly, if we only reduce the kernel size, there may also exist some transitions that introduce extra computational cost by changing the operation type to a more expensive one, \eg, sep\_conv\_$5{\times}5$ $\to$ conv\_$3{\times}3$. Thus, in practice, we make all the transitions meet the above two rules simultaneously to avoid increasing the computational cost. With the proposed two-level transition rule, unlike \sexyname, our \sexynameii is able to automatically obtain the valid transitions for all the operations.}

    \subsubsection{Search Space of \sexynameii}
    \sexynameii has more possible transitions than \sexynamei and thus has a larger search space.
    Given a cell structure with $|\mV|$ nodes and $2(|\mV| {\small -} 3)$ edges, we consider 13 operations/states
    in total (See more details in Fig.~\ref{fig:operation_cost} and supplementary).
	Based on a specific $\beta$, the size of the largest search space of \sexynameii is $| \Omega_{\beta} | = 13^{2(|\mV|-3)}$, which is larger than the largest search space of \sexynamei with the size of $| \Omega_{\beta} | = 3^{2(|\mV|-3)}$.
	Therefore, \sexynameii 
	has the ability to find the architectures with better performance and lower computational cost than \sexynamei (See results in Section~\ref{sec:exp}).
Note that \sexynameii also allows the transitions $O {\to} S$, $O {\to} N$, and $S {\leftrightarrow} N$. Hence, the search space of \sexynamei is a true subset of the search space of \sexynameii.

    \subsubsection{Complexity Analysis of Different Operations}\label{sec:operation_cost}

    Note that our \sexynamei and \sexynameii seek to replace operations with the more efficient ones to avoid introducing additional computation cost. To determine which operations are more efficient,
    we compare the computational cost of different operations in terms of the number of multiply-adds (MAdds) and the number of parameters.

    In Fig.~\ref{fig:operation_cost}, we sort the operations according to the number of parameters and MAdds in descending order.
    {From Fig.~\ref{fig:operation_cost}, we draw the following observations. First, given a fixed kernel size, different operation types have different computational cost. Specifically, separable and dilated separable convolution have lower computational cost than the standard convolution. The max/average pooling, skip connection, and null connection have less or even no computational cost. Second, when we fix the operation type, the kernel size is also an important factor that affects the computational cost of operations. In general, a smaller kernel tends to have a lower computational cost.}

    \subsection{Policy Learning for \sexynameii }\label{sec:policy_natv2}

    To learn the optimal policy $\pi(\cdot|\beta;\theta)$ for \sexynameii, we also use a GCN based classifier to predict the optimal transition for each operation/edge.
    However, it is hard to directly apply the GCN based classifier in \sexyname to predict the optimal transitions for the operations
    with different valid transitions.
	Note that, in \sexyname, all the operations share the same valid transitions, \ie, remaining unchanged, replacing with null connection, replacing with skip connection. 
	However, in \sexynameii,  each operation has its own valid transitions and these transitions directly determine the considered classes of the GCN based classifier. As a result, we may have to design a GCN classifier for each operation, which, however, is very expensive in practice.

    \qi{To address this issue}, we make the following changes to build the GCN model of \sexynameii.
    First, we increase the number of output channels of the final FC layer to match all the considered operations. In this way, \sexynameii is able to consider more possible transitions than \sexynamei.
    Second, according to the transition scheme in Fig.~\ref{fig:state_transition_natv2}, \lzp{we replace the standard softmax layer in Eqn.~(\ref{eq:gcn_classifier}) with a \textbf{Binary-Masked Softmax (BMSoftmax)} layer to omit all the invalid transitions that violate the two-level transition rule.}
    \lzp{Specifically, given $C$ different operations, we represent the transitions for each operation as a binary mask $\bv \in \mmR^C$ (1 for valid transitions and 0 for invalid transitions). To omit the invalid transitions, \sexynameii only computes the probabilities of all the valid transitions and leaves the probabilities of the invalid ones to be zero.
    \qi{Let $\bu {\in} \mmR^C$ be the predicted logits by \sexynameii over $C$ transitions.}
	We compute the probability for the $i$-th transition by}
	\begin{equation}\label{eq:compare_rule} 
        h(\bu, \bv)_i = \frac{\bv_i \cdot e^{\bu_i}}{\sum_{j=1}^K \bv_j \cdot e^{\bu_j}}.
    \end{equation}
	Based on BMSoftmax, \sexynameii is able to determine the optimal transition for the operations with different valid transitions.


\subsection{Possible Bias Risk of \sexynamei and \sexynameii} \label{sec:convergence}

As shown in Figs.~\ref{fig:state_transition} and~\ref{fig:state_transition_natv2}, both \sexynamei and \sexynameii seek to replace redundant operations with skip connections when optimizing architectures. However, the architectures with more skip connections tend to converge faster than other  architectures~\cite{zela2019understanding,chen2020stabilizing}.
As a result, the competition between skip connections and other operations may easily become unfair~\cite{chu2019fairnas} and mislead the search process. 
Consequently, the NAS methods may
incur a bias towards those architectures which converge faster but may yield poor generalization performance~\cite{shu2019understanding,zela2019understanding,chen2020stabilizing}. More analysis on the bias issue can be found in supplementary.

To address the bias issue,  Zhou \etal introduce a
binary gate to each operation and propose a path-depth-wise regularization method to encourage the gates along the long paths in the supernet~\cite{zhou2020theory}. 
Such a regularization forces NAS methods to explore the architectures with slow convergence speed.
It is worth mentioning that, based on NAT and NAT++, we can alleviate the bias issue without the need for complex regularization.
\guo{
As shown in Algorithm~\ref{alg:training}, unlike ENAS~\cite{pham2018efficient} and DARTS~\cite{liu2018darts}, 
we decouple the supernet training from architecture search by sampling architectures from a uniform distribution $p(\cdot)$ rather than the learned policy $\pi(\cdot|\beta;\theta)$. 
Since all the operations have the same probability to be sampled, 
we provide an equal opportunity to train the architectures with different operations. 
In this sense, we can alleviate the possible bias issue (See results in Section~\ref{sec:bias_problem}).
More critically, our methods are able to find better architectures than the architecture searched by~\cite{zhou2020theory} on ImageNet (See Table~\ref{tab:NAS-based}).
}

    \begin{table*}[h]
		\centering
		\caption{Performance of the optimized architectures obtained by different methods based on hand-crafted architectures. ``/'' denotes the original models that are not changed by architecture optimization methods.
		}
		\resizebox{1.0\textwidth}{!}{
			\begin{tabular}{cccccc|ccccccc}
				\toprule[1pt]
				\multicolumn{6}{c|}{\multirow{1}[0]{*}{CIFAR}} & \multicolumn{6}{c}{\multirow{1}[0]{*}{ImageNet}}\\
				\hline
				\multicolumn{1}{c}{\multirow{2}[0]{*}{Model}}  & \multicolumn{1}{c}{\multirow{2}[0]{*}{Method}} & \multicolumn{1}{c}{\multirow{2}[0]{*}{\#Params (M)}} & \multicolumn{1}{c}{\multirow{2}[0]{*}{\#MAdds (M)}} & 
				\multicolumn{2}{c|}{\multirow{1}[0]{*}{Acc. (\%)}} &
				\multicolumn{1}{c}{\multirow{2}[0]{*}{Model}}  & \multicolumn{1}{c}{\multirow{2}[0]{*}{Method}} & \multicolumn{1}{c}{\multirow{2}[0]{*}{\#Params (M)}} & \multicolumn{1}{c}{\multirow{2}[0]{*}{\#MAdds (M)}} & 
				\multicolumn{2}{c}{Acc. (\%)} \\
				\cline{5-6} \cline{11-12}
				& & & & CIFAR-10 & CIFAR-100 & & & & &  \multicolumn{1}{l}{Top-1} & \multicolumn{1}{l}{Top-5} \\
				\hline
				\multirow{6}[0]{*}{VGG16}  & /    &   15.2    &   313   & 93.56 & 71.83 & \multirow{6}[0]{*}{VGG16}  & / &   \multirow{1}[0]{*}{138}    &  \multirow{1}[0]{*}{15620}     & 71.6 & 90.4 \\
				& NAO\cite{luo2018neural}   &   19.5    &   548   & 95.72 & 74.67  &  &  NAO~\cite{luo2018neural}  & 148 & 18896     & 72.9 & 91.3\\
				& ESNAC~\cite{cao2019learnable} &   14.6    &   \textbf{295}   & 95.26 & 74.43 &  &  ESNAC~\cite{cao2019learnable}  & 133 &  14523    & 73.6 & 91.5 \\
				& APS~\cite{wang2020revisiting} &   15.0    &   305   & 95.53 & 74.79 &  &  APS~\cite{wang2020revisiting}  & 137 &  15220    & 73.9 & 91.7 \\
				& \sexynamei    &   15.2    &   315   & {96.04} & 75.02
				& &  \sexyname &   138   &    15693   & {74.3} & {92.0} & \\
				& \sexynameii & \textbf{14.4} & {301} & \textbf{96.16} & \textbf{75.23} & & \sexynameii & \textbf{131} & \textbf{14907} & \textbf{74.7} & \textbf{92.2} \\
				\hline
				\multirow{6}[0]{*}{ResNet20} & /    &   0.3    &   41   & 91.37 & 68.88 & \multirow{6}[0]{*}{ResNet18} & /  &   \multirow{1}[0]{*}{11.7}     &  \multirow{1}[0]{*}{1580}     & 69.8 & 89.1 \\
				& NAO~\cite{luo2018neural}      &   0.4    &   61   & 92.44  & 71.22 & &  NAO~\cite{luo2018neural}   &   17.9     &  2246     & 70.8 & 89.7\\
				& ESNAC~\cite{cao2019learnable} &   0.3    &  40     & 92.87 & 71.58 &  &  ESNAC~\cite{cao2019learnable}  & 11.2  &  1544    & 71.0 & 89.9 \\
				& APS~\cite{wang2020revisiting} & 0.3      &   42   & 93.14 & 71.84 &  &  APS~\cite{wang2020revisiting}  & 11.2 &  1547    & 70.9 & 90.0 \\
				& \sexynamei     &   0.3    &   42   & {93.05} & 71.67 & &  \sexynamei  &    11.7    &  1588 &  {71.1}  & {90.0}  \\
				& \sexynameii & \textbf{0.3} & \textbf{39} & \textbf{93.23} & \textbf{71.97} & & \sexynameii & \textbf{11.0} & \textbf{1516} & \textbf{71.3} & \textbf{90.2} \\
				\hline
				\multirow{6}[0]{*}{ResNet56} & /      &   0.9    &   127   & 93.21 & 71.54 & \multirow{6}[0]{*}{ResNet50} &   /  &  \multirow{1}[0]{*}{25.6}     &  \multirow{1}[0]{*}{3530}     & 76.2 & 92.9\\
				& NAO~\cite{luo2018neural}       &   1.3    &   199   & 95.27  & 74.25 & &  NAO~\cite{luo2018neural}     &  34.8     &  4505     & 77.4 & 93.2 \\
				& ESNAC~\cite{cao2019learnable} &   0.8    &  125    & 95.33 & 74.30 &  &  ESNAC~\cite{cao2019learnable}  & 25.0 &   3484    & 77.4 & 93.3 \\
				& APS~\cite{wang2020revisiting} &   0.8    &  \textbf{123}   & 94.54 & 73.58 &  &  APS~\cite{wang2020revisiting}  & 24.9  &  3461    & 77.6 & 93.4 \\
				& \sexynamei     &   0.9    &   129   & {95.40} & 74.33 &  &  \sexynamei &  25.6    &   3547    & {77.7} & {93.5} \\
				& \sexynameii & \textbf{0.8} & {124} & \textbf{95.47} & \textbf{74.41} & & \sexynameii & \textbf{24.8} & \textbf{3452} & \textbf{77.8} & \textbf{93.6} \\
				\hline
				\multirow{6}[0]{*}{ShuffleNet} & /      &   0.9    &  161   & 92.29 & 71.14 & \multirow{6}[0]{*}{ShuffleNet} &   /  &  \multirow{1}[0]{*}{2.4}     &  \multirow{1}[0]{*}{138}     & 68.0 & 86.4 \\
				& NAO~\cite{luo2018neural}       &    1.4   &   251   &  93.16 & 72.04 & &  NAO~\cite{luo2018neural}     &   3.5    &   217    & 68.2 & 86.5 \\
				& ESNAC~\cite{cao2019learnable} & 0.8 &  153    &  93.21 & 72.14 &  &  ESNAC~\cite{cao2019learnable}  & 2.2 & 131 & 68.4 & 86.6\\
				& APS~\cite{wang2020revisiting} &   0.9    &  161  &   {93.47} & 72.40 & &  APS~\cite{wang2020revisiting}  & 2.4 & 138  & \textbf{68.9}  & 87.0 \\
				& \sexynamei     &   0.8    &  158    & {93.37} & 72.34  &   &  \sexynamei &   2.3   &    136   & 68.7 & 86.8 \\
				& \sexynameii & \textbf{0.7} & \textbf{147} & \textbf{93.53} & \textbf{72.61} & & \sexynameii & \textbf{2.1} & \textbf{125} & {68.8}  & \textbf{87.0} \\
				\hline
				\multirow{5}[0]{*}{MobileNetV2} & /      &   {2.3}    &  {91}   & 94.47 & 73.66 & \multirow{5}[0]{*}{MobileNetV2} &   /  &  \multirow{1}[0]{*}{{3.4}}     &  \multirow{1}[0]{*}{{300}}     & 72.0 & 90.3 \\
				& NAO~\cite{luo2018neural}       &   2.9    &   131   & 94.75  & 73.79 & &  NAO~\cite{luo2018neural}     &  4.5     &  513     & 72.2 & 90.6 \\
				& ESNAC~\cite{cao2019learnable} &   \textbf{2.1}    & \textbf{84} & 94.87  & 73.94 &  &  ESNAC~\cite{cao2019learnable}  & \textbf{3.1} &   \textbf{277}   & 72.4 & 90.8\\
				& APS~\cite{wang2020revisiting} &  2.3      &   90   & 95.03 & 74.14 &  &  APS~\cite{wang2020revisiting}  & 3.4  &  303    & 72.3 & 90.6 \\
				& \sexynamei /\sexynameii     &   {2.3}    &   92   & \textbf{95.17} & \textbf{74.22} & &  \sexynamei/\sexynameii  &  {3.4}    &   302    & \textbf{72.5} & \textbf{91.0} \\
				\bottomrule[1pt]
			\end{tabular}
		}
		\label{tab:hand-crafted}
	\end{table*}
	
	\begin{table}[t]
		\centering
		\caption{Performance comparisons of the optimization results of hand-crafted architectures on face recognition datasets. ``/'' denotes the original models that are not changed by architecture optimization methods.
		}
		\resizebox{1.0\linewidth}{!}
		{
		\LARGE
			\begin{tabular}{ccccccc}
				\toprule[1pt]				\multicolumn{1}{c}{\multirow{2}[0]{*}{Model}}  & \multicolumn{1}{c}{\multirow{2}[0]{*}{Method}} & \multicolumn{1}{c}{\multirow{2}[0]{*}{\#Params (M)}} & \multicolumn{1}{c}{\multirow{2}[0]{*}{\#MAdds (M)}} & \multicolumn{3}{c}{\multirow{1}[0]{*}{Acc. (\%)}}\\
				\cline{5-7}
				& & & & LFW & CFP-FP & AgeDB-30 \\
				\hline
				\multirow{6}[0]{*}{LResNet34E-IR~\cite{deng2019arcface}}  & /    &   31.8    &   7104   & 99.72 & 96.39 & 98.03 \\
				& NAO\cite{luo2018neural}   &   43.7    &  9874    &  99.73 & 96.41 & 98.07   \\
				& ESNAC~\cite{cao2019learnable}   &   31.7   &  7002    &  99.77 & 96.52 & 98.19   \\
				& APS~\cite{wang2020revisiting}   &   31.6   &  {6997}    &  99.80 & 96.64 & 98.30   \\
				& \sexynamei    &   31.8    &   7107   & 99.79 & 96.66 & 98.28 \\
				& \sexynameii & {31.5} & {7023} & \textbf{99.83} & \textbf{96.72} & \textbf{98.35}    \\
				\hline
				\multirow{5}[0]{*}{MobileFaceNet~\cite{chen2018mobilefacenets}} & /    &   1.0    &   441   & 99.50 & 92.23 & 95.63 \\
				& NAO~\cite{luo2018neural}      &  1.3     &  584    &  99.53 & 92.28 & 95.75 \\
				& ESNAC~\cite{cao2019learnable}   &   {0.9}    &  {408}    &  99.59 & 92.37 & 95.98  \\
				& APS~\cite{wang2020revisiting}   &   1.0    &  437    &  99.63 & 92.41 & 96.13    \\
				& \sexynamei/\sexynameii     &    1.0   & 443  & \textbf{99.76} & \textbf{92.50} & \textbf{96.36}    \\
				\bottomrule[1pt]
			\end{tabular}
		}
		\label{tab:face}
	\end{table}

\section{Experiments}\label{sec:exp}
	
	We apply our method to optimize some well-designed architectures, including hand-crafted architectures and NAS based architectures.
    We have released the code for both \sexynamei\footnote{The code of \sexynamei is available at \href{https://github.com/guoyongcs/NAT}{https://github.com/guoyongcs/NAT}.} and \sexynameii\footnote{The code of \sexynameii is available at \href{https://github.com/guoyongcs/NATv2}{https://github.com/guoyongcs/NATv2}.}.

	\begin{figure*}[t!]
		\centering
		\includegraphics[width=0.86\linewidth]{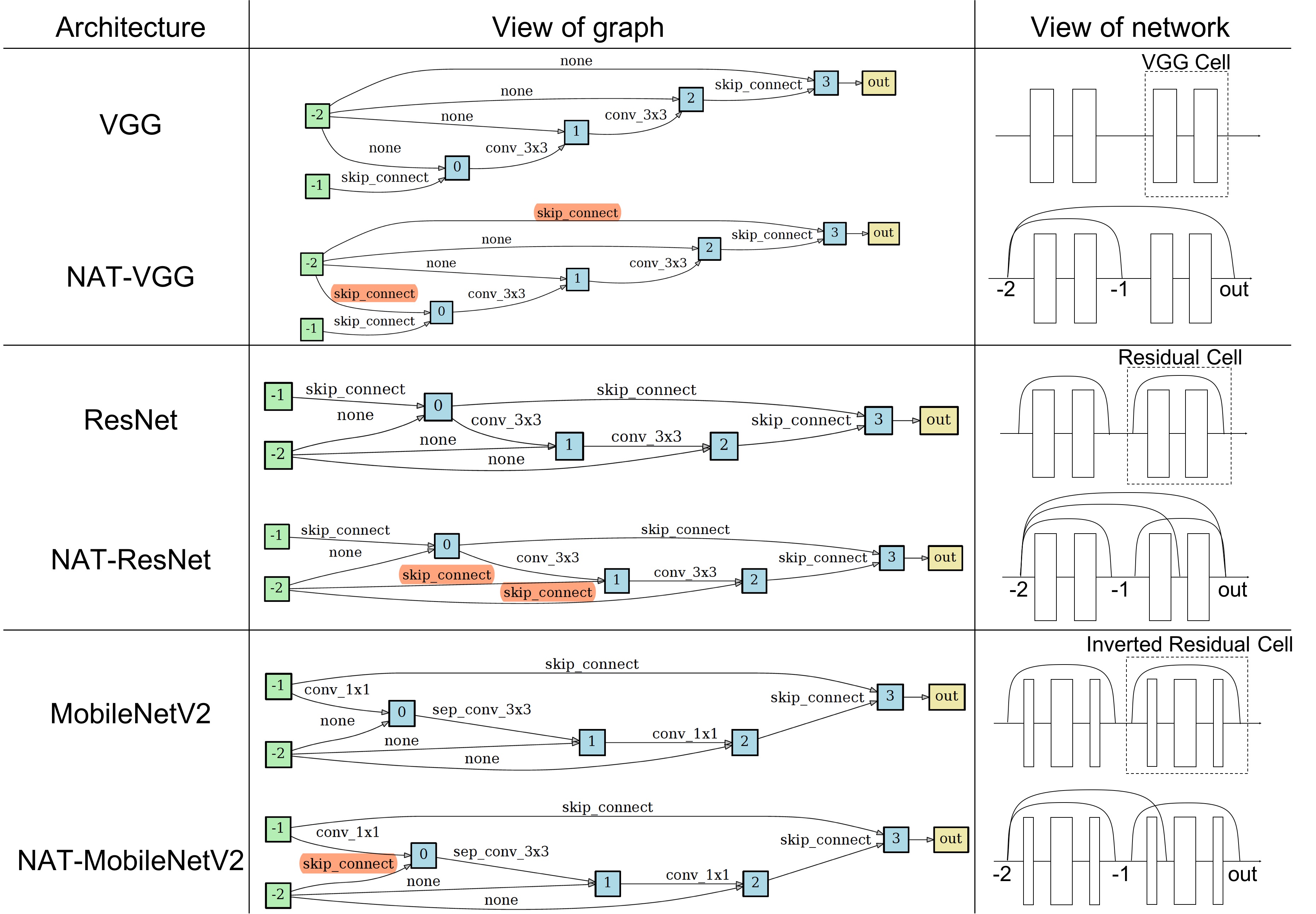}
		\caption{Architecture optimization results of several hand-crafted architectures. We provide both the views of graph (middle) and network (right).
		}
		\label{fig:vgg_resnet}
	\end{figure*}

	\subsection{Implementation Details}\label{sec:implementation}

    	\begin{table*}[t]
		\centering	\caption{Comparisons of the optimized architectures obtained by different methods on NAS based architectures. ``-'' denotes that the results are not reported. ``/'' denotes the original models that are not changed by architecture optimization methods. $^\dagger$ denotes the models trained with cutout.
		}
		\resizebox{1.0\textwidth}{!}{
			\begin{tabular}{cccccc|ccccccc}
				\toprule[1pt]
				\multicolumn{6}{c|}{\multirow{1}[0]{*}{CIFAR}} & \multicolumn{6}{c}{\multirow{1}[0]{*}{ImageNet}}\\
				\hline
				\multicolumn{1}{c}{\multirow{2}[0]{*}{Model}}  & \multicolumn{1}{c}{\multirow{2}[0]{*}{Method}} & \multicolumn{1}{c}{\multirow{2}[0]{*}{\#Params (M)}} & \multicolumn{1}{c}{\multirow{2}[0]{*}{\#MAdds (M)}} & 
				\multicolumn{2}{c|}{\multirow{1}[0]{*}{Acc. (\%)}} &
				\multicolumn{1}{c}{\multirow{2}[0]{*}{Model}}  & \multicolumn{1}{c}{\multirow{2}[0]{*}{Method}} & \multicolumn{1}{c}{\multirow{2}[0]{*}{\#Params (M)}} & \multicolumn{1}{c}{\multirow{2}[0]{*}{\#MAdds (M)}} & 
				\multicolumn{2}{c}{Acc. (\%)} \\
				\cline{5-6} \cline{11-12}
				& & & & CIFAR-10 & CIFAR-100 & & & & & \multicolumn{1}{l}{Top-1} & \multicolumn{1}{l}{Top-5} \\
				\hline
				AmoebaNet$^\dagger$~\cite{real2018regularized} & \multirow{4}[0]{*}{/} & 3.2 & - &  96.73 & - & AmoebaNet~\cite{real2018regularized} & \multirow{4}[0]{*}{/} & 5.1 & 555 &  74.5 & 92.0 \\
				PNAS$^\dagger$~\cite{liu2018progressive} &  & 3.2 & - & 96.67 & 81.13 & PNAS~\cite{liu2018progressive} &  & 5.1 & 588 &  74.2 & 91.9 \\
				SNAS$^\dagger$~\cite{xie2018snas} &  & 2.9 & - & 97.08 & 82.47 & SNAS~\cite{xie2018snas} &  & 4.3 & 522 &  72.7 & 90.8 \\
				GHN$^\dagger$~\cite{zhang2018graph} &  & 5.7 & - & 97.22 & - & GHN~\cite{zhang2018graph} &  & 6.1 & 569 &  73.0 & 91.3 \\
				
				PR-DARTS$^\dagger$~\cite{zhou2020theory} &  & 3.4 & - & 97.68 & 83.55 & PR-DARTS~\cite{zhou2020theory} &  & 5.0 & 543 &  75.9 & 92.7 \\
				\hline
                \multirow{6}[0]{*}{ENAS$^\dagger$~\cite{pham2018efficient}}  & /     &   4.6    &   804   & 97.11 & 82.87 & \multirow{6}[0]{*}{ENAS~\cite{pham2018efficient}} & /  &    5.6   &   607    & 73.8  & 91.7 \\
                & NAO~\cite{luo2018neural}      &   4.5    &   763   & 97.05  & 82.57 & &  NAO~\cite{luo2018neural}    &    5.5   &   589   & 73.7 & 91.7 \\
                & ESNAC~\cite{cao2019learnable}      &   4.1    &    717  & 97.13  & 83.15 & &  ESNAC~\cite{cao2019learnable}    &  \textbf{5.0}     &   \textbf{542}   & 73.5 & 91.4 \\
                & APS~\cite{wang2020revisiting}      &   4.4    &  744    & 97.26  & 83.45 & &  APS~\cite{wang2020revisiting}    &   5.5    &  591    & 74.0 & 91.9 \\
                & \sexynamei     &  4.6    &    804   & {97.24} & 83.43 &  &  \sexynamei & 5.6   &   607    & {73.9} & {91.8} \\
				& \sexynameii    & \textbf{3.7} & \textbf{580} & \textbf{97.31} & \textbf{83.51} & & \sexynameii & {5.4} & {582} & \textbf{74.3} & \textbf{92.1} \\
				\hline
                \multirow{6}[0]{*}{DARTS$^\dagger$~\cite{liu2018darts}}  & /   &   3.3    &   533   & 97.06 & 83.03 & \multirow{6}[0]{*}{DARTS~\cite{liu2018darts}} & /  &    4.7   &   574    & 73.1 & 91.0\\
                & NAO~\cite{luo2018neural}   &   3.5    &   577   & 97.09 & 83.12 & &  NAO~\cite{luo2018neural}     &    5.0   &   621    & 73.3 & 91.1  \\
                & ESNAC~\cite{cao2019learnable}      &  2.8     &    457  & 97.21  & 83.36 & &  ESNAC~\cite{cao2019learnable}    &    4.0   &    494  & 73.5 & 91.2 \\
                & APS~\cite{wang2020revisiting}      &   3.2    &   515  & 97.25  & 83.44 & &  APS~\cite{wang2020revisiting}    &    4.5   & 539  & 73.3  & 91.2  \\
                & \sexynamei    &   {2.7}    &   {424}    &  {97.28} & 83.49 & & \sexynamei  & {4.0}    &   {441}    & {73.7} & {91.4} \\
				& \sexynameii    & \textbf{2.5} & \textbf{395} & \textbf{97.30} & \textbf{83.56} & & \sexynameii & \textbf{3.8} & \textbf{413} & \textbf{73.9} & \textbf{91.5} \\
				\hline
				\multirow{6}[0]{*}{NAONet$^\dagger$~\cite{luo2018neural}}  & /   &   128    &   66016   & 97.89 & 84.33 & \multirow{6}[0]{*}{NAONet~\cite{luo2018neural}} & /  &    11.3   &   1360    & 74.3 & 91.8 \\
				& NAO~\cite{luo2018neural}   &   143    &  73705    & 97.91 & 84.42 & &  NAO~\cite{luo2018neural}     &  11.8     &  1417     & 74.5  & 92.0  \\
                & ESNAC~\cite{cao2019learnable}      &    107   & 55187      &  97.98 & 84.49 & &  ESNAC~\cite{cao2019learnable}    &   9.5    &   1139   & 74.6 & 92.1 \\
                & APS~\cite{wang2020revisiting}      &   125    &   63468   &  97.96 & 84.47 & &  APS~\cite{wang2020revisiting}    &    11.0   &    1286  & 74.5 & 92.1  \\
				& \sexynamei    &         113    & 58326  & {98.01} & 84.53 & & \sexynamei  &  8.4     &  1025     & {74.8} & {92.3} \\
				& \sexynameii    &  \textbf{101} & \textbf{51976} & \textbf{98.07} & \textbf{84.60} & & \sexynameii    & \textbf{8.1} & \textbf{992} & \textbf{75.0} & \textbf{92.5} \\
				\hline
				\multirow{6}[0]{*}{PC-DARTS$^\dagger$~\cite{xu2020pcdarts}}  & /   &   3.6    &   570   & 97.43 & 84.21  & \multirow{6}[0]{*}{PC-DARTS~\cite{xu2020pcdarts}} & /  &    5.3   &   597    & 75.8 & 92.7  \\
				& NAO~\cite{luo2018neural}   &   4.7     &   725   & 97.49 & 84.30 & &  NAO~\cite{luo2018neural}     &   6.7    &   706    & 76.0  & 92.8   \\
                & ESNAC~\cite{cao2019learnable}      &   {3.3}    & \textbf{503}  & 97.44  & 84.20 & &  ESNAC~\cite{cao2019learnable}    &   \textbf{4.7}    &    \textbf{529}  & 75.9 & 92.7  \\
                & APS~\cite{wang2020revisiting}      &   3.4    &   529   & 97.47  & 84.28 & &  APS~\cite{wang2020revisiting}    &   5.0    &  557   & 76.0 & 92.7 \\
				& \sexynamei    &       3.4      & 518  & 97.51  & 84.31 & & \sexynamei  &    4.9   &   546    & 76.1  & 92.8  \\
				& \sexynameii    &  \textbf{3.3} & 512 & \textbf{97.57} & \textbf{84.37} & & \sexynameii    & 4.8 & 540 & \textbf{76.3} & \textbf{93.0} \\
				\bottomrule[1pt]
			\end{tabular}
		}
		\label{tab:NAS-based}%
	\end{table*}

	{We build the supernet by stacking 8 cells with the initial channel number of 20. We train the transformer for 200 epochs.} 
	Following the setting of~\cite{liu2018darts}, we set $m=1$, $n=1$, and $\lambda=0.003$ in Eqn.~(\ref{eq:entropy_gradient}).
	To cover all possible architectures, we set $q(\cdot)$ to be a uniform distribution.
	{For} the evaluation of networks, 
	we replace the original cells with the optimized {cells} and train the models from scratch. 
	For all the considered architectures, we follow the same settings of the original papers, \ie, we build the models with the same number of layers and channels as the original ones. 
	We only apply cutout to the NAS based architectures on CIFAR.

	\subsection{Results on Hand-crafted Architectures} \label{sec:results_hand_crafted}

	\lzp{In this experiment, we apply both \sexyname and \sexynameii to four popular hand-crafted architectures, namely VGG~\cite{simonyan2014very}, ResNet~\cite{he2016deep}, ShuffleNet~\cite{zhang2018shufflenet} and MobileNetV2~\cite{sandler2018mobilenetv2}.}
	To make all architectures share the same graph representation method defined in Section~\ref{sec:MDP}, we add null connections into the hand-crafted architectures to ensure that each node has two input nodes (See Fig.~\ref{fig:vgg_resnet}).
	Note that each hand-crafted architecture may have multiple graph representations. However, our methods yield stable results on different graph representations (See results in supplementary).

\subsubsection{Quantitative Results}

From Table~\ref{tab:hand-crafted},
our \sexynamei based models consistently outperform the original models by a large margin with approximately the same computational cost.
Compared to \sexynamei, \sexynameii produces better optimized architectures with higher accuracy and lower computational cost. These results show that, by enlarging the search space, \sexynameii is able to {further improve the performance of architecture optimization.}
\lzp{Moreover, compared to existing methods (\ie, NAO, ESNAC and ASP), NAT++ produces the architectures with higher accuracy and lower computational cost.}
Note that \sexynamei and \sexynameii yield the same results when optimizing MobileNetV2. The main reason is that the operations in MobileNetV2 are either conv\_$1{\times}1$ or sep\_conv\_$3{\times}3$, which have already been very efficient operations. Thus, it is hard to benefit from the extended transition scheme of \sexynameii when there are very few valid operation transitions.

	\lzp{We also evaluate our method on face recognition tasks. In this experiment, we consider three benchmark datasets (\ie, LFW~\cite{huang2007labeled}, CFP-FP~\cite{sengupta2016frontal} and AgeDB-30~\cite{moschoglou2017agedb}) and two baselines (\ie, LResNet34E-IR~\cite{deng2019arcface} and MobileFaceNet~\cite{chen2018mobilefacenets}). 
	We adopt the same settings as that in~\cite{deng2019arcface}. 
	More training details can be found in the supplementary.
    From Table~\ref{tab:face}, 
	the models optimized by \sexyname consistently outperform the original models without introducing extra computational cost. 
	Moreover, \sexynameii yields the best optimization results \wrt both architectures on all datasets. 
	}
	
	\subsubsection{Visualization of the Optimized Architectures}

	In this section, we visualize the original and optimized hand-crafted architectures in Fig.~\ref{fig:vgg_resnet}.
	From Fig.~\ref{fig:vgg_resnet},
	\sexynamei is able to introduce additional skip connections to the architecture to improve the architecture design. 
	Unlike \sexynamei, \sexynameii
	conducts architecture optimization in a finer manner.
	Specifically, \sexynameii replaces some standard convolutions with separable convolutions for VGG and ResNet. In this way, \sexynameii not only reduces the number of parameters and computational cost but also further improves the performance (See Table~\ref{tab:hand-crafted}).

	\subsection{Results on NAS Based Architectures}\label{exp:results_nas}

    {We also apply the proposed methods to}
	the automatically searched architectures.
	In this experiment, we consider \lzp{four state-of-the-art NAS based architectures, namely DARTS~\cite{liu2018darts}, ENAS~\cite{pham2018efficient}, NAONet~\cite{luo2018neural} and PC-DARTS~\cite{xu2020pcdarts}.}
	Moreover, we compare our optimized architectures with other NAS based architectures, including AmoebaNet~\cite{real2018regularized}, PNAS~\cite{liu2018progressive}, SNAS~\cite{xie2018snas}, 
	GHN~\cite{zhang2018graph}, 
    and PR-DARTS~\cite{zhou2020theory}.

   \begin{figure}[t!]
		\centering
        	\centering
        	\includegraphics[trim = 12mm 0mm 20mm 5mm,
		clip, width=1.0\columnwidth]{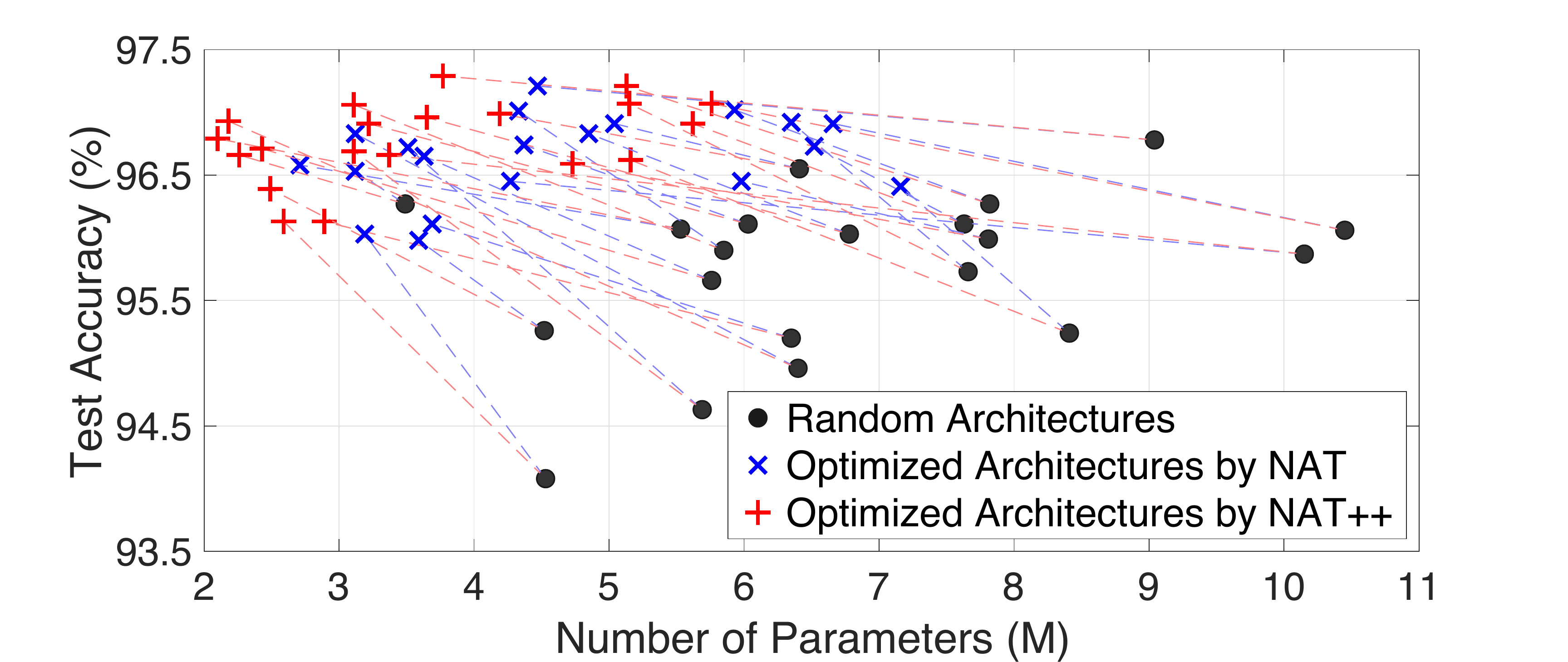}
        	\label{fig:random_arch_fig}
        \vspace{-10pt}
        \caption{
        Effect of \sexyname and \sexynameii on the average performance over 20 randomly sampled architectures on CIFAR-10.
        }
        \label{fig:random_arch_fig}
    \end{figure}

	From Table~\ref{tab:NAS-based}, {given different input architectures,}
	{the architectures obtained by \sexynamei consistently
	yield higher accuracy than their original counterparts and the architectures optimized by existing methods.}
	For example, given DARTS as input, \sexynamei not only reduces 15\% parameters and 23\% computational cost but also achieves 0.6\% improvement in terms of Top-1 accuracy on ImageNet. For NAONet, \sexynamei reduces approximately 25\% parameters and computational cost, and achieves 0.5\% improvement in terms of Top-1 accuracy.
    Moreover, we also evaluate the architectures optimized by \sexynameii. As shown in Table~\ref{tab:NAS-based}, {equipped with the extended transition scheme,} \sexynameii is able to find better architectures with higher accuracy and lower computational cost than the architectures found by \sexynamei \lzp{and existing methods}.
    \guo{Due to the page limit, we show the visualization results of the optimized architectures in the supplementary.}
    These results show the effectiveness of the proposed method.

\section{Further Experiments}

\begin{figure}
    \centering
    \subfigure[Number of layers vs. Accuracy.]{
        \includegraphics[width=0.49\columnwidth]{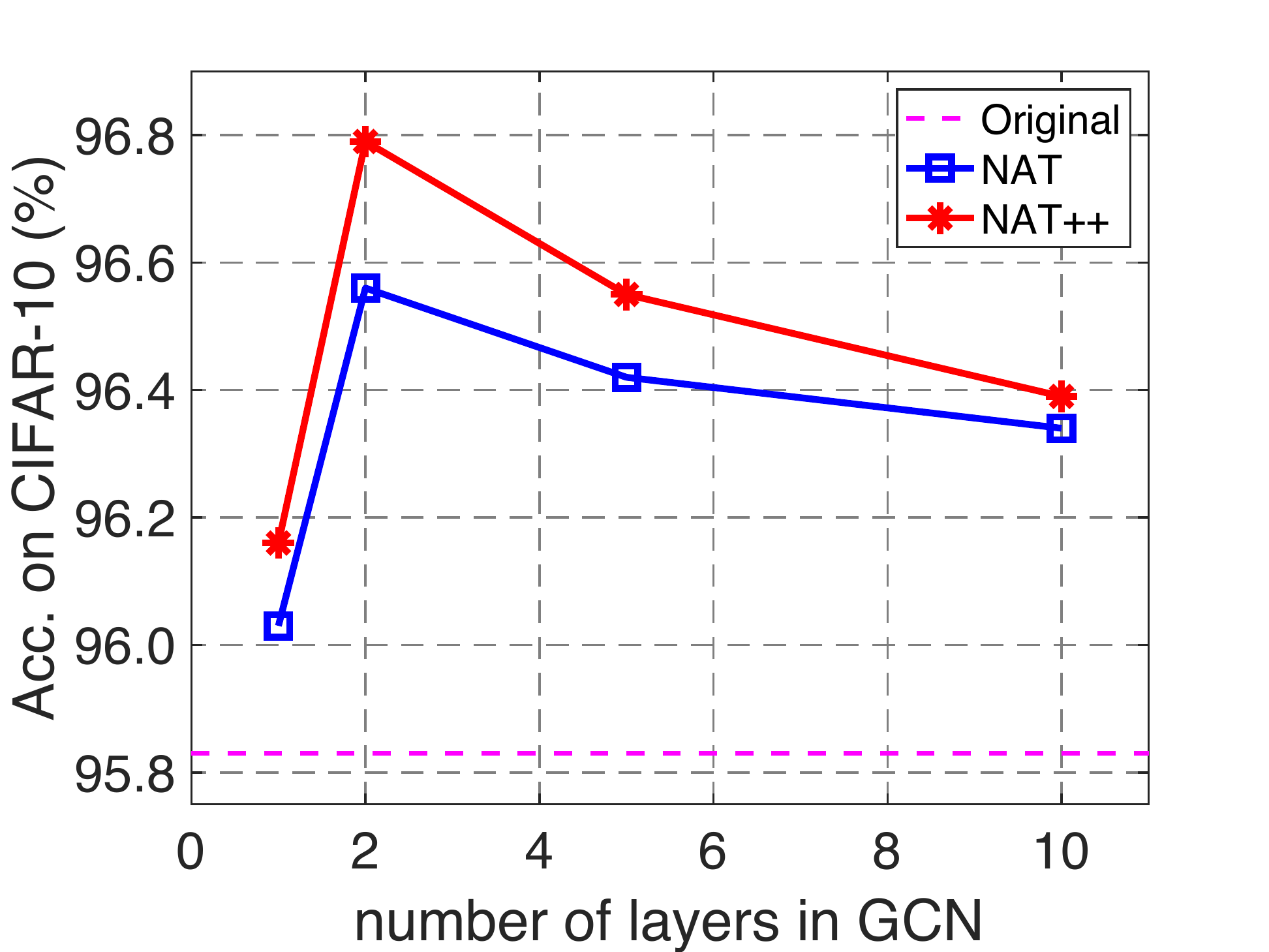}
        \label{fig:num_gcn}
    }~
    \subfigure[Value of $\lambda$ vs. Accuracy.]{
        \includegraphics[width=0.49\columnwidth]{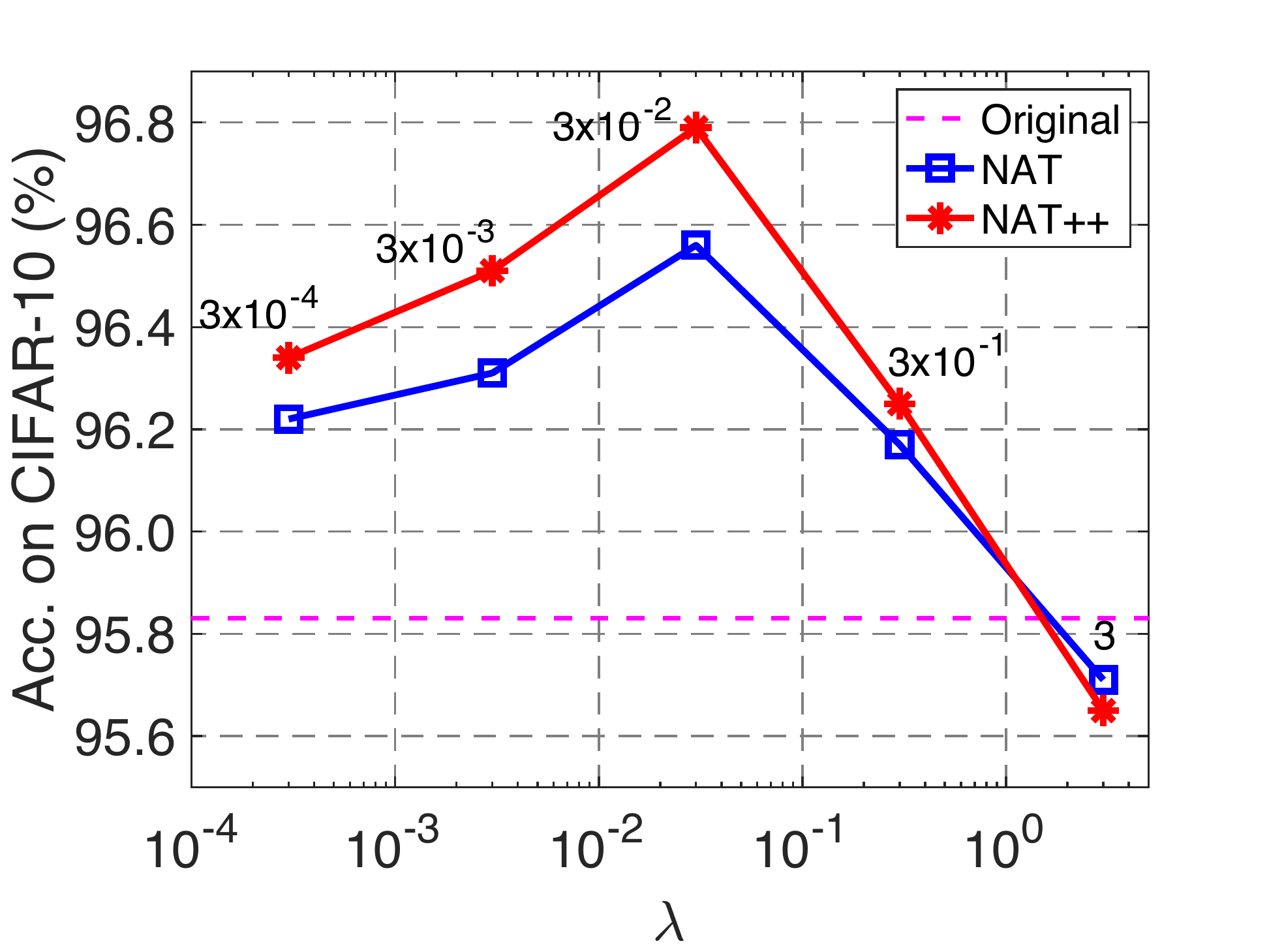}
        \label{fig:lambda}
    }
    \caption{Effect of the number of layers in GCN and the value of $\lambda$ on the performance of \sexynamei and \sexynameii.}
\end{figure}

	\begin{figure*}[t!]
    \centering
	\subfigure[Convergence curves of Arch1.]{
		\includegraphics[width = 0.48\columnwidth]{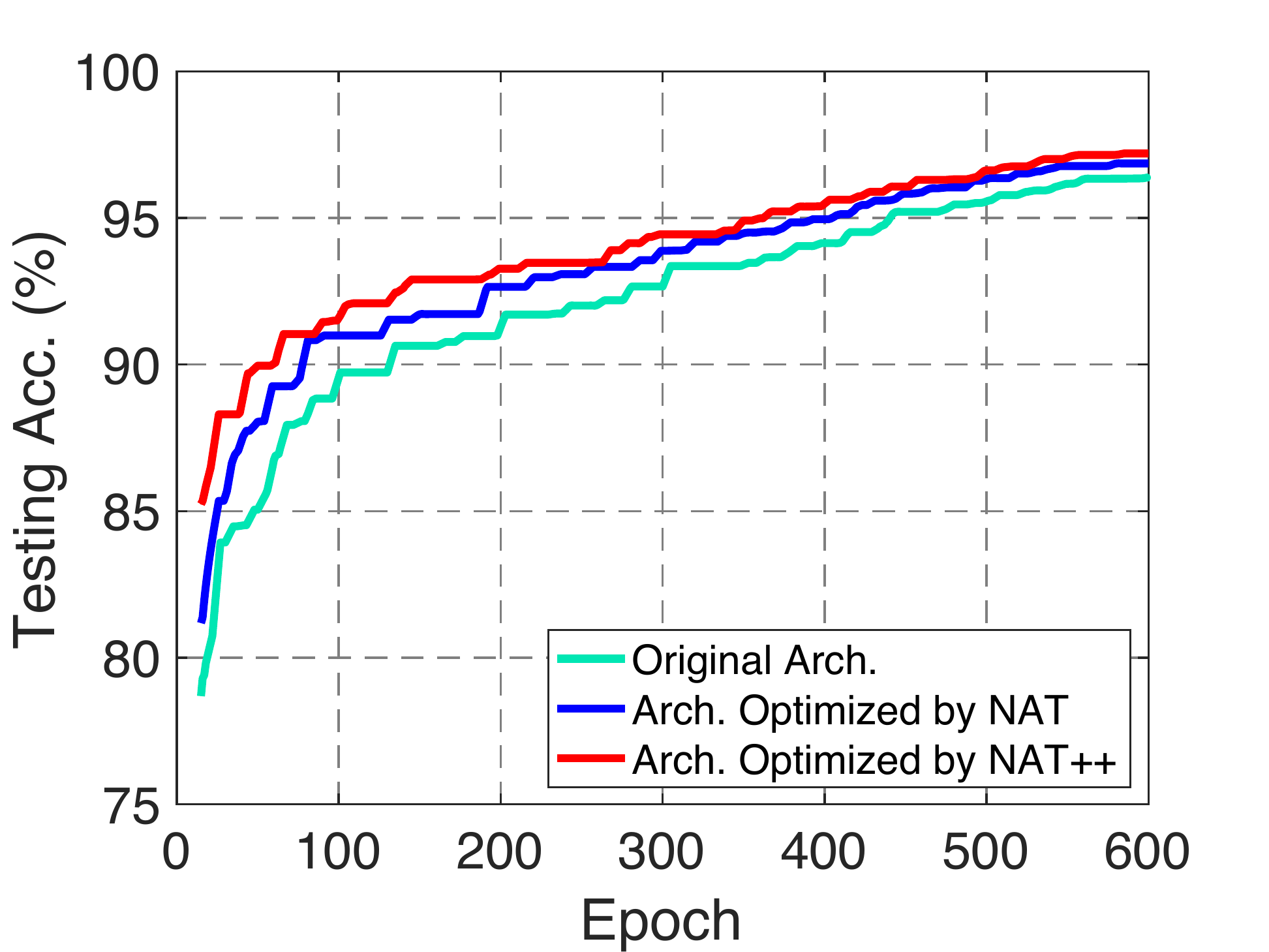}\label{fig:arch1}
	}~
	\subfigure[Convergence curves of Arch2.]{
		\includegraphics[width = 0.48\columnwidth]{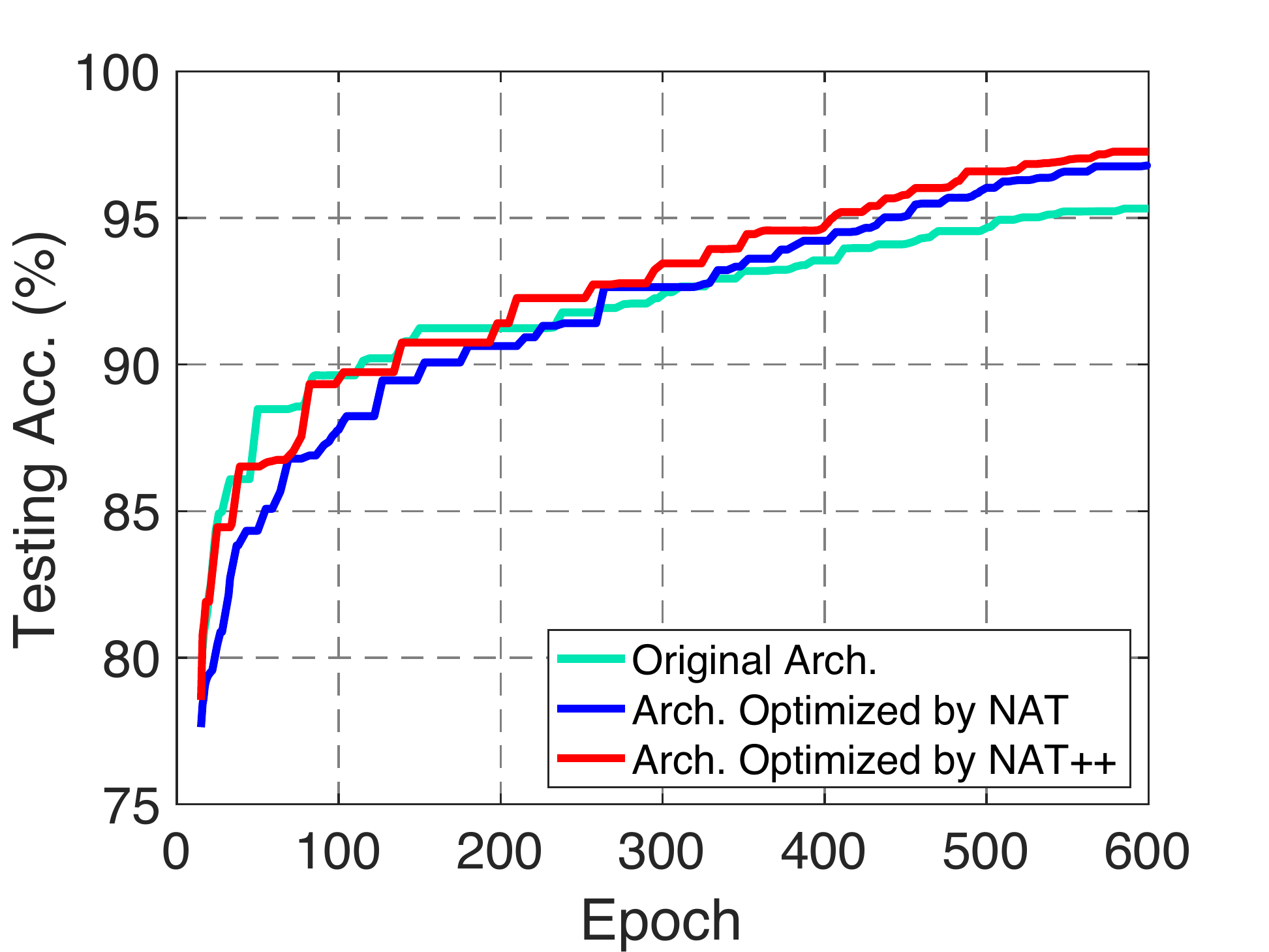}\label{fig:arch1}
	}~
	\subfigure[Convergence curves of Arch3.]{
		\includegraphics[width = 0.48\columnwidth]{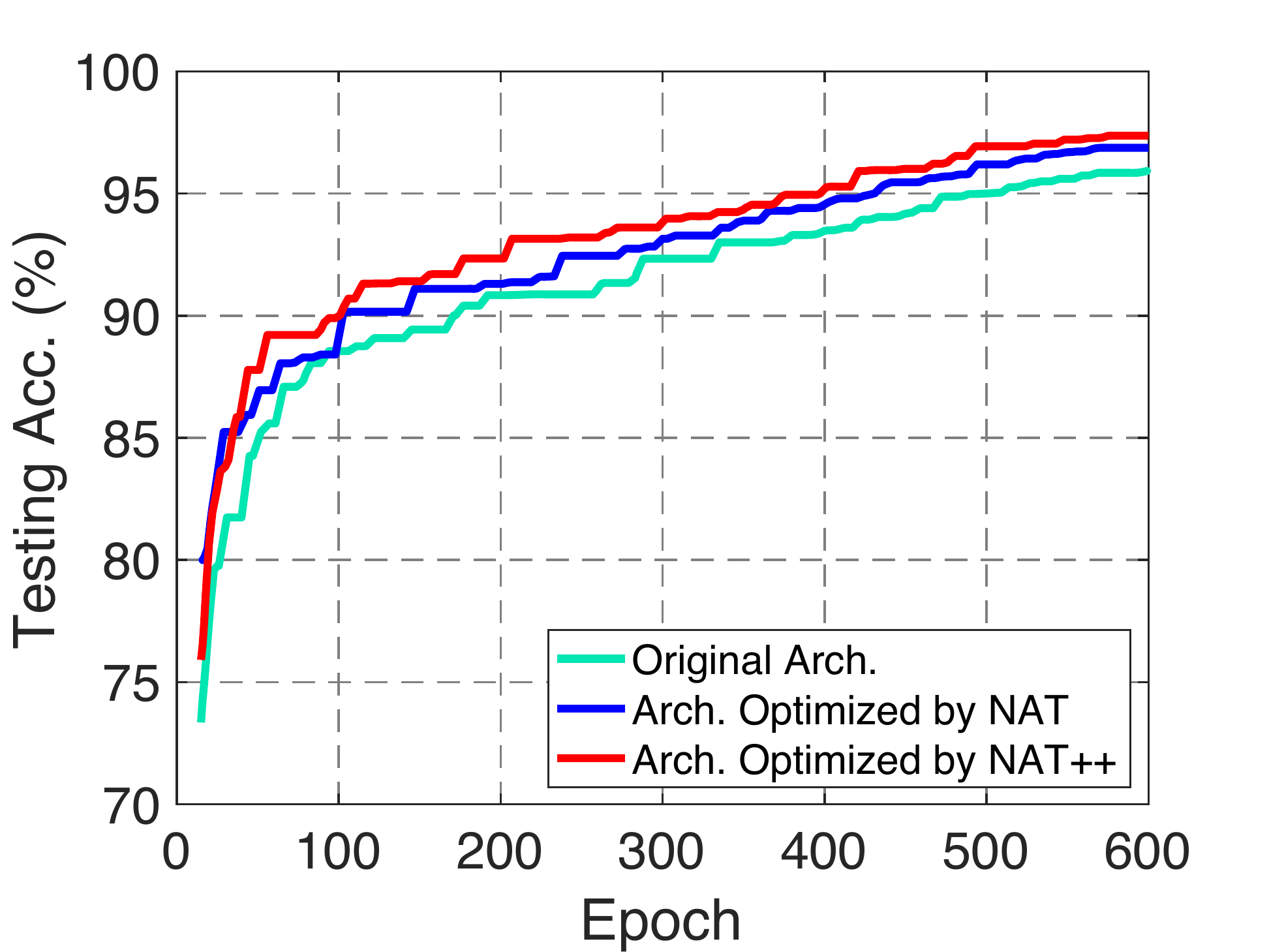}\label{fig:arch1}
	}~
	\subfigure[Convergence curves of Arch4.]{
		\includegraphics[width = 0.48\columnwidth]{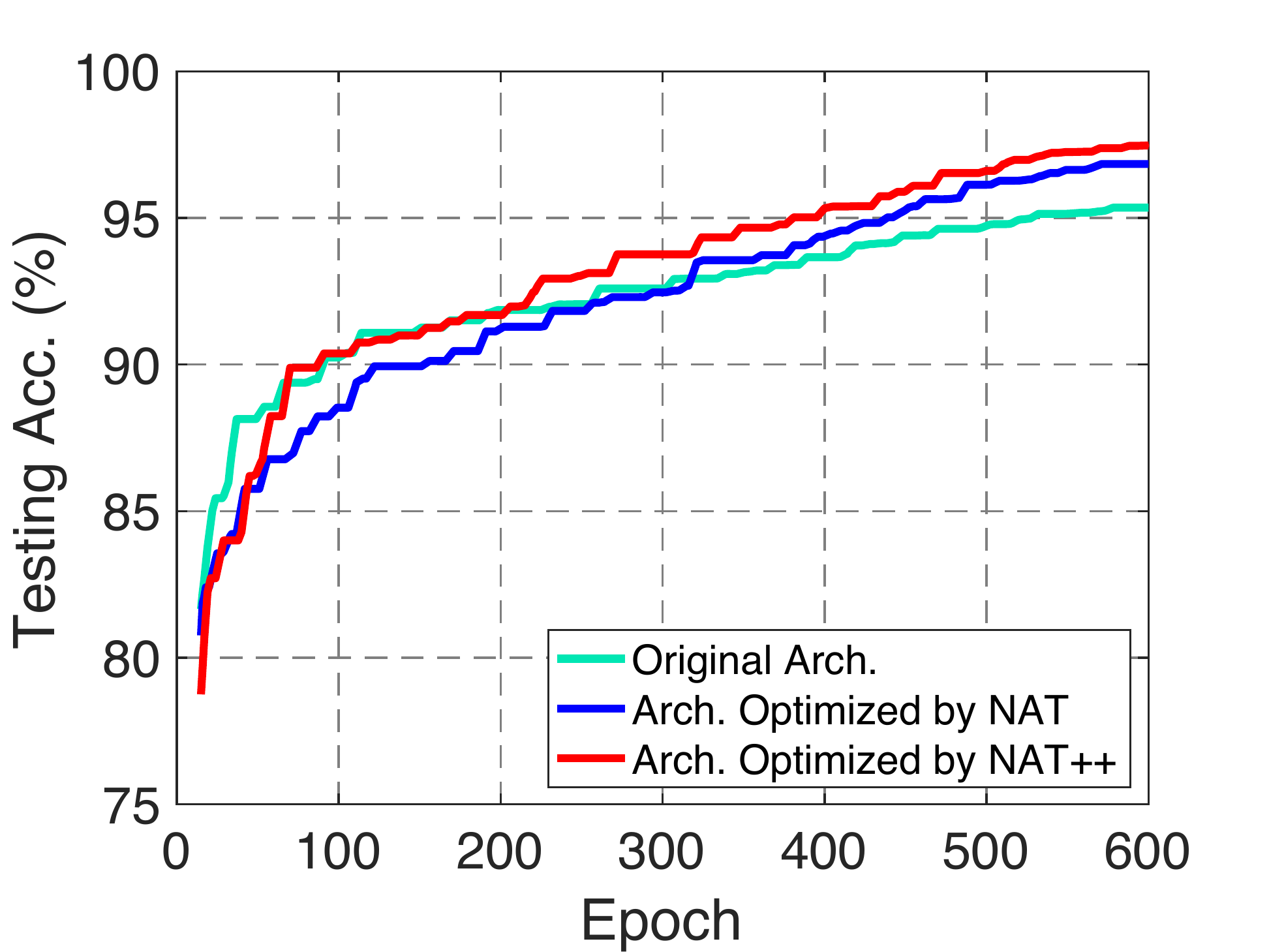}\label{fig:arch1}
	}
    \caption{Comparisons of convergence between the original architectures and the optimized architectures by \sexynamei and \sexynameii on CIFAR-10.}
    \label{fig:convergence}
    \end{figure*}

	\subsection{Results on Randomly Sampled Architectures} \label{sec:random_sample_arch}
    We apply our \sexynamei and \sexynameii to 20 randomly sampled architectures from the whole architecture space. 
	We train all architectures using momentum SGD with {a} batch size of 128 for 600 epochs.
	From Table~\ref{tab:random_arch_tab} and Fig.~\ref{fig:random_arch_fig},
	the architectures optimized by \sexynamei surpass the original ones in terms of both accuracy and computational cost. 
	Moreover, equipped with the two-level transition scheme, \sexynameii further improves the architecture optimization results. 
	To better illustrate this, we exhibit the result of each architecture in Fig.~\ref{fig:random_arch_fig}, which shows that the models optimized by \sexynameii achieve higher accuracy with fewer parameters than \sexynamei.
	In this sense, our method has good generalizability on a wide range of architectures, making it possible to be applied in real-world applications.

	\begin{table}[t!]
		\centering
		\caption{Effect of \sexyname on the average performance over 20 randomly sampled architectures on CIFAR-10. 
		}
		\small
		\resizebox{0.75\linewidth}{!}
		{
            \begin{tabular}{c|c|c|c}
                \toprule[1pt]
                \multicolumn{1}{c|}{\multirow{1}[0]{*}{Method}} & \multicolumn{1}{c|}{\multirow{1}[0]{*}{Original}} & \multicolumn{1}{c|}{\sexynamei} & \multicolumn{1}{c}{\sexynameii}  \\
                \hline
                \#Params (M)  & 6.40$\pm$2.04 & 4.67$\pm$1.36 & {\bf 3.66$\pm$1.23} \\
                \#MAdds (G)  & 1.07$\pm$0.32  & 0.79$\pm$0.21 & {\bf 0.52$\pm$0.20} \\
                Test accuracy (\%)  & 95.83$\pm$1.08  & 96.56$\pm$0.47 & {\bf 96.79$\pm$0.32} \\
                \bottomrule[1pt]
            \end{tabular}%
        }
      \label{tab:random_arch_tab}
    \end{table}

\subsection{Effect of the Number of Layers in GCN}
\lzp{We investigate the effect of the number of layers in GCN on the performance of our method. Specifically, we apply both \sexynamei and \sexynameii to optimize 20 randomly sampled architectures.
We build 4 GCN models with $\{1,2,5,10\}$ layers, respectively. 
Note that a graph convolutional layer aims to extract features by aggregating the information from the neighbors of each node (\ie, one-hop neighbors)~\cite{wu2020comprehensive}. The GCN with multiple layers is able to exploit the information from multi-hop neighbors in a graph~\cite{lin2018multi,ding2019cognitive}.
}

\lzp{
From Fig.~\ref{fig:num_gcn}, when we build a single-layer GCN, the model yields very poor performance since a single-layer model cannot handle the information from the nodes with more than 1 hop. However, if we build the GCN model with 5 or 10 layers, the larger models also hamper the performance since the models with too many graph convolutional layers (\ie, high-order model) may introduce redundant information~\cite{zhu2019multi}. 
To learn a good policy, we build a two-layer GCN in practice.}

\subsection{Effect of $\lambda$ in Eqn.~(\ref{eq:obj_entropy})} 

\lzp{In this part, we investigate the effect of $\lambda$ (which makes a trade-off between the reward and the entropy term in Eqn.~(\ref{eq:obj_entropy})) on the performance of architecture optimization. 
We train \sexynamei and \sexynameii with $\lambda \in \{3 {\times} 10^{-4}, 3 {\times} 10^{-3}, 3 {\times} 10^{-2}, 3 {\times} 10^{-1}, 3\}$ and report the average accuracy over the optimization results of 20 randomly sampled architectures.
}
\lzp{
From Fig.~\ref{fig:lambda}, when we increase $\lambda$ from $3 {\times} 10^{-4}$ to $3 {\times} 10^{-2}$, the entropy term gradually becomes more important and encourages the model to explore the search space. In this way, it prevents the model from converging to a local optimum and helps find better optimized architectures. If we further increase $\lambda$ to $3 \times 10^{-1}$, the entropy term would overwhelm the objective function and hamper the performance. When we use a very large $\lambda=3$, the search process becomes approximately the same as random search and yields the architectures even worse than the original counterparts. In practice, we set $\lambda = 3 {\times} 10^{-2}$.
}

\begin{table}[t]
  \centering
  \caption{Effect of $m$ and $n$ on the performance and search cost (GPU hour) of \sexyname and \sexynameii. We report the average performance on CIFAR-10 over the optimization results of 20 randomly sampled architectures.}
  \resizebox{1.0\linewidth}{!}
  {
  \large
    \begin{tabular}{c|c|cccc}
    \hline
    \multicolumn{2}{c|}{m} & 1        & 5     & 10    & 30 \\
    \hline
    \multirow{2}[0]{*}{NAT} & Accuracy (\%)  &   96.56$\pm$0.47       &    96.58$\pm$0.41   &   \textbf{96.61$\pm$0.33}    & 96.59$\pm$0.37 \\
          & Search Cost & \textbf{5.3}    & 19.9 & 38.9 & 122.7 \\
    \hline
    \multirow{2}[0]{*}{NAT++} & Accuracy (\%)   &   96.79$\pm$0.32    &      96.80$\pm$0.37      &   \textbf{96.83$\pm$0.29}    &  96.81$\pm$0.34 \\
          & Search Cost  & \textbf{5.7}    & 20.8 & 40.4 & 114.3 \\
    \hline
    \hline
    \multicolumn{2}{c|}{n} & 1     & 5     & 10    & 30 \\
    \hline
    \multirow{2}[0]{*}{NAT} & Accuracy (\%)  &  96.56$\pm$0.47        &   \textbf{96.59$\pm$0.41}    &    96.57$\pm$0.43   &  96.58$\pm$0.39 \\
          & Search Cost  & \textbf{5.3}   & 17.1 & 33.3 & 82.2 \\
    \hline
    \multirow{2}[0]{*}{NAT++} & Accuracy (\%)   &   96.79$\pm$0.32    &   96.80$\pm$0.35         &   96.82$\pm$0.35    &  \textbf{96.84$\pm$0.37} \\
          & Search Cost  & \textbf{5.7}    & 18.2 & 35.1 & 86.7 \\
    \hline
    \end{tabular}%
    }
  \label{tab:effect_m_n}%
\end{table}%

\subsection{Effect of $m$ and $n$ in Eqn.~(\ref{eq:entropy_gradient})}

\lzp{In this section, we investigate the effect of the hyper-parameters $m$ and $n$ on the performance of our method.
When we gradually increase $m$ during training, more architectures have to be evaluated via additional forward propagations through the supernet to compute the reward.
The search cost would significantly increase with the increase of $m$. From Table~\ref{tab:effect_m_n}, we do not observe obvious performance improvement when we consider a large $m$. 
One possible reason is that, based on the uniform distribution $p(\cdot)$, even sampling one architecture in each iteration has provided sufficient diversity of the input architectures to train our model.
Thus, 
we set $m=1$ in practice.}

\lzp{We also investigate the effect of the hyper-parameter $n$ which controls the number of sampled optimized architectures for each input architecture. 
When we consider a large $n$, we have to evaluate more optimized architectures to compute the reward in each iteration, yielding significantly increased search cost. 
As shown in Table~\ref{tab:effect_m_n}, similar to $m$, our model only achieves marginal performance improvement with the increase of $n$.
In practice, $n=1$ works well in \sexynamei and \sexynameii.
The main reason is that most of the sampled architectures can be very similar based on a fixed policy/distribution $\pi(\cdot|\beta;\theta)$. As a result, increasing the number of sampled optimized architectures may provide limited benefits for the training process. Actually, a similar phenomenon is also observed in ENAS~\cite{pham2018efficient}.}

\subsection{Discussions on the Possible Bias Risk} \label{sec:bias_problem}
\guo{
In this section, based on our methods, we investigate the possible bias issue towards the architectures that have fast convergence speed (in the early stage) but poor generalization performance.
In this experiment, we randomly collect a set of architectures and use \sexynamei and \sexynameii to optimize them. Then, we compare the convergence curves of the original architectures and the optimized architectures on CIFAR-10. 
From Fig.~\ref{fig:convergence}, some of the original architectures incur the issue of ``fast convergence in the early stage but with poor generalization performance'', \eg, Arch2 and Arch4. In contrast, all of the architectures optimized by \sexynamei and \sexynameii have a relatively stable convergence speed and yield better generalization performance than their original counterparts. 
From these results, the bias problem is not obvious in our methods.
The main reason is that, in \sexynamei and \sexynameii, all the operations have the same probability to be sampled and we would offer an equal opportunity to train the architectures with different operations. 
In this sense, we are able to alleviate the too fast convergence issue incurred by skip connection.
Due to the page limit, we put the convergence curves of more architectures in the supplementary.}

	\section{Conclusion}
	
	In this paper, we {have proposed} a novel Neural Architecture {Transformer} (\sexyname) {for the task of architecture optimization}. To solve this problem,  
	{we seek to replace the existing operations with more computationally efficient operations.
	Specifically, we propose a \sexynamei to replace the redundant or non-significant operations with the skip connection or null connection. Moreover, we design an advanced \sexynameii to further enlarge the search space. \guo{To be specific, we present a two-level transition rule which encourages operations to have a more efficient type or smaller kernel size to produce the more compact architectures.}}
	To verify the proposed method, we apply {\sexynamei and \sexynameii} to optimize both hand-crafted architectures and Neural Architecture Search (NAS) based architectures. Extensive experiments on \lzp{several benchmark} datasets demonstrate the effectiveness of the proposed method in improving the accuracy and compactness of neural architectures.
	

	
	\appendices

	\ifCLASSOPTIONcaptionsoff
	\newpage
	\fi

	

	
	

	\bibliographystyle{IEEEtran}
	\bibliography{nat}

\end{document}